\documentclass[amsfonts, amssymb, amsmath, showkeys, twocolumn]{revtex4-2}

\usepackage{graphicx}
\usepackage{amsmath}
\usepackage{array}
\usepackage{amssymb}
\usepackage{color}
\usepackage{float}
\usepackage{dcolumn}
\usepackage{bm}

\usepackage{mathabx}
\usepackage{enumitem} 
\usepackage{tabularx}
\usepackage{soul}
\usepackage[dvipsnames]{xcolor}

\usepackage{diagbox}
\usepackage{easyReview}

\begin{document}
\title{
{Computational imaging with the human brain}}

\author{G. Wang$^{1}$, D. Faccio$^{1}$}
    \email[Correspondence email address: ]{daniele.faccio@glasgow.ac.uk}
    \affiliation{School of Physics \& Astronomy, University of Glasgow, G12 8QQ Glasgow, UK}

\date{\today} 

\begin{abstract}
{Brain-computer interfaces (BCIs) are enabling a range of new possibilities and routes for augmenting human capability. Here, we propose BCIs as a route towards forms of computation, i.e. computational imaging, that blend the brain with external silicon processing.  
We demonstrate ghost imaging of a hidden scene using the human visual system that is combined with an adaptive computational imaging scheme. 
This is achieved through a projection pattern `carving' technique that relies on real-time feedback from the brain to modify patterns at the light projector, thus enabling more efficient and higher resolution imaging. This brain-computer connectivity demonstrates a form of augmented human computation  that could in the future extend the sensing range of human vision and provide new approaches to the study of the neurophysics of human perception. { {As an example, we illustrate a simple experiment whereby image reconstruction quality is affected by simultaneous conscious processing and readout of the perceived light intensities.}}}
\end{abstract}


\maketitle

{\bf{Introduction.}} 
Neurotechnologies and specifically brain-computer interfaces (BCIs) provide a route to augmenting human cognitive abilities, with applications ranging from decision making to memory enhancement \cite{cinel,raisamo,lelievre2013single,vourvopoulos2012robot,van2013experiencing,bi2013eeg,wang2011decoding,brumberg2010brain,prataksita2014brain,abdulkader2015brain}.
Visual control of BCIs is a specific example of interface that typically relies on the so-called steady state visual evoked potential (SSVEP) and that can be read-out either with implanted electrodes or more readily, using an electroencephalogram (EEG) \cite{WOLPAW201367,wang2008brain,norciaSteadystateVisualEvoked2015,vialatteSteadystateVisuallyEvoked2010}. In this case, it is the visual system that acts as sensor of the surrounding environment and/or controller of the computer. SSVEP requires a periodically repeating illumination pattern  or light modulation, typically in the 3-4 Hz up to 30-40 Hz region, to stimulate a steady-state (periodic) response in the brain. 
A well-known feature of SSVEP is also the strong nonlinearity in the form of multiple harmonics in the output power spectrum \cite{tuncelModelBasedInvestigation2019,kimDifferentialRolesFrequencyfollowing2011,labeckiNonlinearOriginSSVEP2016}.\\
A question that we address here, builds upon BCIs based on visually evoked responses in the brain and relates to  whether the brain can  be integrated into forms of computational imaging.\\
Computational imaging is the use of computer-based approaches to complement or enhance machine vision or imaging. Notable examples relevant to this work include ghost imaging, i.e. image formation with just one single detector pixel and non-line-of-sight imaging, i.e. the ability to see behind corners. \\
In its simplest version, ghost imaging (GI) relies on illuminating an object with a series of light patterns and then detecting only the corresponding reflected or transmitted gray-scale intensity values that will vary due to the different spatial overlap of each pattern with the object. By weighting each pattern with the corresponding measured intensity value and summing over all patterns, one can reconstruct an image of the object. {Significant research has been devoted also to the problem of optimising the shape or required number of illumination patterns, including also compressive sensing techniques} \cite{zhaoGhostImagingLidar2012,pittmanOpticalImagingMeans1995,zhangCorrelatedTwophotonImaging2005,pellicciaExperimentalXrayGhost2016,xu1000FpsComputational2018,zhao2019ultrahigh,katkovnikCompressiveSensingComputational2012,sun2012normalized,yaoIterativeDenoisingGhost2014,heGhostImagingBased2018,shimobabaComputationalGhostImaging2018,moreauGhostImagingUsing2018a,padgett2017introduction,padgett2016ghost,shapiroComputationalGhostImaging2008}. \\
Non-line-of-sight imaging instead, refers to the ability to reconstruct images of scenes that are hidden from sight by e.g. a wall or obstacle. The more common approaches to this involve using a pulsed light source to illuminate the hidden scene. The reflected light is collected in the form of transient waveforms that are reflected from a secondary surface, e.g. another wall and are recorded with very high (i.e. picosecond) temporal resolution. Various inverse retrieval or processing techniques including artificial neural networks (ANNs) can then retrieve the final image. GI protocols have also been implemented for forms of NLOS imaging that rely on the fact the GI only requires collecting an intensity value that is retro-reflected from the hidden scene image, as long as one knows the exact shape of the illumination patterns \cite{katkovnikCompressiveSensingComputational2012,baiImagingCornersSinglepixel2017,faccioNonlineofsightImaging2020,saundersComputationalPeriscopyOrdinary2019,otooleConfocalNonlineofsightImaging2018,veltenRecoveringThreedimensionalShape2012,chenLearnedFeatureEmbeddings2020}.\\
A key point of these and other computational techniques is that they rely on some form of machine-based detection, i.e. cameras or single pixel sensors and these are then combined with computational algorithmic approaches to retrieve scene images.\\
In this work we propose a route towards brain-computer forms of computational imaging. 
We demonstrate a ghost imaging protocol that relies on relaying light intensity information reflected from a surface and that is read-out as an SSVEP from the brain. This information is then processed by a computer-based algorithm and an artificial neural network that reconstructs an image from the SSVEP power spectrum. { {This imaging process is made more efficient by an adaptive computational loop whereby the SSVEP signal also indicates how to select the appropriate illumination patterns that are sent on to the scene being imaged. We then show preliminary results whereby the reconstructed imaging quality is used to quantify the difference between nonconscious processing of the light intensity (through the EEG signal) and explicit conscious processing (by asking the participant to either verbally communicate or type on a keyboard the perceived light intensity)}}. \\
%
{\bf{Imaging protocol.}} 
Computational ghost imaging relies on a  light source that can project a series of typically binary (black and white) patterns, $P_n$. These light patterns are then reflected (or transmitted) from the object or scene we wish to image and collected with a bucket detector (i.e. sensitive only to total energy), $a_n$. Then, summation of all the bucket value-weighted patterns will produce an image, $O=\sum a_{n}P_{n}$. A very common choice of patterns are the Hadamard set, $H$, that can be recursively defined.\\
Over the years, researchers have optimised ghost imaging by using different light sources, detectors or computational algorithms. 
Recent attempts have also used the human visual system as a detector where the visual persistence time of the retina is used to directly perform the summation operation described above, i.e. a series of pre-weighted patterns are pre-calculated and are then visualised at sufficiently high rates that they are effectively perceived by the eye as an accumulated sum \cite{boccoliniGhostImagingHuman2019,wangAllOpticalNakedEyeGhost2020,wangNakedeyeGhostImaging2020}.
{ {Conversely, here we implement a form of computational ghost imaging in which the human visual cortex processes visual data and also provides feedback that allows to adapt the projected patterns in real-time so as to minimise measurement time.}}  \\
\begin{figure}[t]%
\centering
\includegraphics[width=\columnwidth]{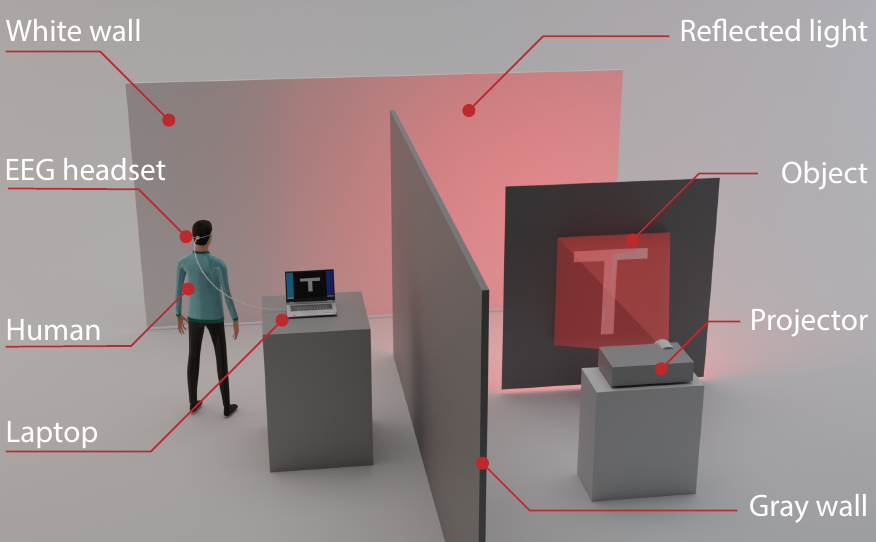}
\caption{The setup used for { {adaptive}} ghost imaging. A light projector illuminates an object cut out from a cardboard support. Transmitted light is diffused by a ground glass that is in contact with the cardboard support and illuminates a white, observation wall. This part of the setup is obscured from the observer by a wall. The distance of both the object and the observer from this secondary wall is $\sim0.5-1$ m. The EEG signal from the observer is recorded and processed on a computer.}\label{GI_sys}
\end{figure}
 {\bf{Ghost imaging with the brain. }} A schematic overview of the experiments is shown in Fig.~\ref{GI_sys}. We project a series of binary Hadamard patterns using a standard digital light projector (DLP) onto an object. The light transmitted past the object is then observed in reflection from a secondary white surface (white wall).  Each binary pattern is periodically switched on/off for several periods with a frame rate that chosen in the 3-30 Hz region. We detect the SSVEP generated by the  visual cortex activity from a single electrode placed at $O_z$, the medial visual cortex region (see SI). This SSVEP is then analysed in the spectral domain and the corresponding fundamental (i.e. at the same frequency of the light modulation) and higher harmonic (due to neuron nonlinearity) amplitudes are extracted. These are then used to reconstruct an image of the object, which as shown in the schematic overview, is hidden behind a wall.  \\
{\bf{Linearity. }} The first step for any form of imaging requires calibration of the detection system and identification of linear regions or at least regions in which the system response is monotonic with increasing input intensity. In this case, the `system' is the visual system and SSVEP read-out, which is known to exhibit significant nonlinearity. We characterised the (non)linearity of the SSVEP readout with a standard LCD screen that displayed a flickering uniform intensity with frequency between 3 and 30 Hz and that was varied across the full 8 bit range of the screen, i.e. in values from 0 to 255, corresponding to completely black (no light) and very bright (corresponding to 125 Lumens).
The EEG signal is then Fourier transformed \cite{auger1995improving,fulop2006algorithms}. Clear harmonic peaks are observed as expected \cite{norciaSteadystateVisualEvoked2015} and we then consider the  maximum values
of the individual harmonics (up to the fourth) as well as the total sum of these values (the total SSVEP energy).  The SSVEP energy heatmap for each individual harmonic shows a complicated and typically non-monotonic dependence for varying screen intensity and flicker frequency (see Supplementary Information).\\
Figure~\ref{I_vs_H_curve}a shows the total SSVEP energy. Here we can identify two ideal flicker frequency regions at 6 and 15 Hz, shown in  Fig.~\ref{I_vs_H_curve}b. The region around 15 Hz shows a clear monotonic increase of SSVEP energy with increasing illumination and a similar behaviour occurs also at 6 Hz, albeit only for a more limited screen intensity range (between 0 and $\sim125$ bits, i.e. between 0 and $\sim75$ Lumens). The same calibration measurements performed across three different people resulted in a similar behaviour (see Supplementary Information). We therefore perform most of our experiments at either 15 Hz (using the full 0-125 Lumens intensity range) or 6 Hz (using a limited intensity range).\\
\begin{figure}[t]%
\centering
\includegraphics[width=\columnwidth]{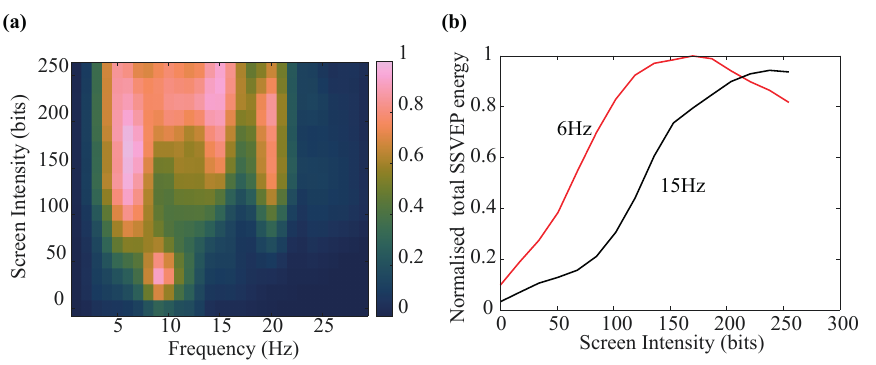}
\caption{{\bf{(a)}} Heatmap of the measured total SSVEP energy (sum of all harmonic peaks). {\bf{b}} Total SSVEP energy at 6 Hz and 15 Hz. 
}\label{I_vs_H_curve}
\end{figure}
{\bf{Ghost Imaging results.}} 
Using the setup shown in Fig.~\ref{GI_sys}, objects are illuminated with Hadamard patterns that are each periodically flickered  (see SI for full details).\\ 
\begin{figure}[t]%
\centering
\includegraphics[width=\columnwidth]{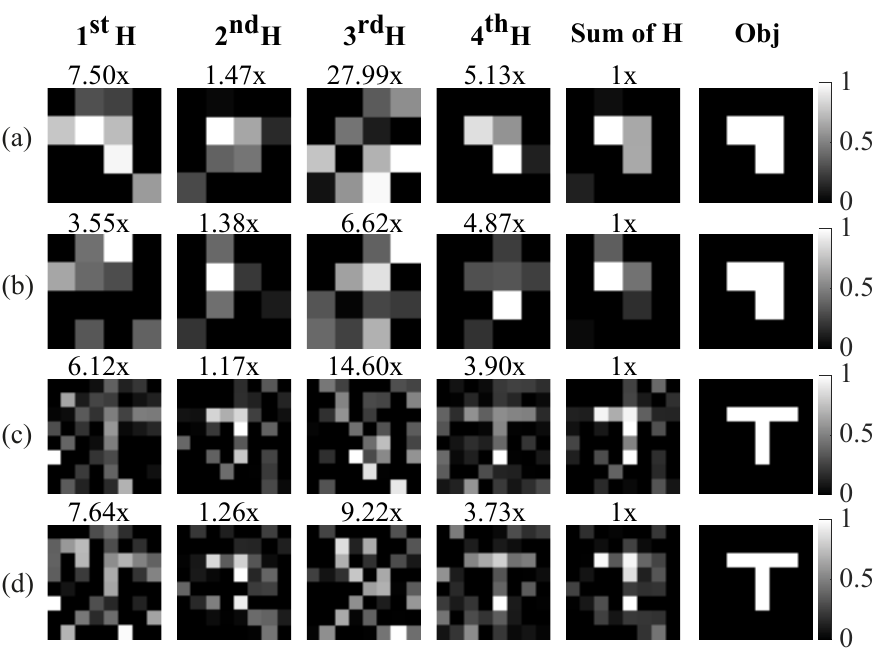}
\caption{Standard ghost imaging results. (a) Inverted ``L'' shape (4 sec/pattern illumination time; { {total acquisition (illumination) time of 84 seconds}}). (b) Inverted ``L'' shape (2 sec/pattern illumination time; { {total acquisition  time of 42 seconds}}). (c) Letter ``T'' (8 sec illumination time; { {total acquisition  time of 512 seconds}}). (d) Letter ``T'' (4 sec illumination time; { {total acquisition time of 256 seconds}}). The columns, from left to right, show the ghost images that are reconstructed from the SSVEP fundamental ($1^{st}$H), second harmonic ($2^{nd}$H), third harmonic ($3^{rd}$H), fourth harmonic ($4^{th}$H) and total energy (sum over all 4 harmonics). The last column shows the ground truth object shape. Each image is normalised to the `Sum of harmonics' total intensity (rescaling factors are shown above each image).\\
}\label{GI_result_4_4s}
\end{figure}
Figures~\ref{GI_result_4_4s}(a) and (b) show results for the standard ghost imaging approach for a $4 \times 4$ pixel object with a 6 Hz flicker frequency and for 4 s and 2 s illumination time for each of the first 16 Hadamard patterns. The columns show the ghost image reconstruction obtained using each individual harmonic SSVEP energy and then for the total energy (sum over all harmonics). Only the total SSVEP energy allows to reconstruct a clear image, in keeping with the calibration tests. More complicated images require more pixels. For example, Figs.~\ref{GI_result_4_4s}(c) and (d) show the attempts to image the letter ``T'' on an $8 \times 8$ pixel grid. At 4 s illumination time (Fig.~\ref{GI_result_4_4s}(d)), we obtain only a very noisy image. Increasing the illumination time to 8 s for each pattern (Fig.~\ref{GI_result_4_4s}(c)), provides a marginally better image where the letter ``T'' is starting to emerge, hinting that significantly increasing illumination times could lead to better images. However, this strategy would lead to impractical experiment times that could then lead to other problems, including fatigue for the viewer. \\
%
%
%
%
{\bf{{ {Adaptive}} ghost imaging with the human brain.}}  An adaptive  feedback loop is employed to adjust the projected Hadamard patterns dynamically during the measurement process and thus improve both the imaging speed and the image quality. 
The underlying principal of this is a `Hadamard matrix carving' method that is based on the observation that when projecting Hadamard (or any given choice of) patterns onto an object, not all patterns will have significant overlap with the object and this can be used to dynamically adapt the choice of successive projections.\\%
In brief, patterns are taken from the Hadamard matrix $H$. This matrix has columns composed of vector Hadamard patterns, each of length equal to the total number of pixels in the image, $N$, and therefore, $H$ has  rank $N$.  These patterns (i.e. columns taken from $H$) are projected one at a time. Whenever a bucket value is measured that is below a certain threshold, this indicates that this specific pattern has a minimal or zero overlap with the object. We therefore apply a `row carving' operator, $R$, that `carves' the Hadamard matrix by removing all rows corresponding to the non-zero row elements of the pattern. The resulting matrix will have a reduced rank, $N/2$. We then apply a `column carving' operator, $C$, that removes columns that do not contribute to increasing the matrix rank. In this way we obtain a new square, carved matrix $H_c=RHC$ that also has rank $N/2$. This process is then repeated on $H'$, with additional carving being applied each time a pattern is found with no overlap with the object, each time reducing by a factor 2x the rank and therefore the number of required illumination patterns. The final result will be a reduced $H_c$ that contains $N/2^m$ patterns instead of $N$ with a corresponding reduction in measurement time. The precise value of $m$ and therefore of the reduction of the measurement time, depends on the specific details of the object that is being imaged. In general terms, sparse binary objects can lead to very significant gains in terms of patterns that are dropped with a significant decrease of measurement time, as shown below.\\
Full details with a worked example of the Hadamard carving approach are provided in the Supplementary material.\\
%
%
{\bf{Image reconstruction.}}
Various approaches can be implemented to reconstruct the final image. As seen above, the  standard GI where the image is reconstructed as $O=\sum a_nH_n$, will give rather noisy images. \\
We can use the carving approach described above and then reconstruct an image from $O=(H_c\cdot H_c^T)^{-1}\cdot H_c\cdot B$, where $B$ indicates the vector formed by all the measured SSVEP values (see SI for details). We can additionally use the patterns that were eliminated as masks that indicate where we should expect the image to have zero intensity. 
This `carved ghost imaging' (CGI) approach leads to significant improvement by removing noise from pixels outside the object.\\
Finally, we implemented an end-to-end deep neural network (DNN-GI) that consists of both an image reconstruction and denoising step. The DNN is therefore composed of a linear layer that is trained to reconstruct an image taking as an input only the detected bucket (i.e. SSVEP) values followed by a series of
nonlinear layers for denoising \cite{zhang2017beyond} (see details in SI).\\
%
%
%
%
%
\begin{figure}[t]%
	\centering
	\includegraphics[width=\columnwidth]{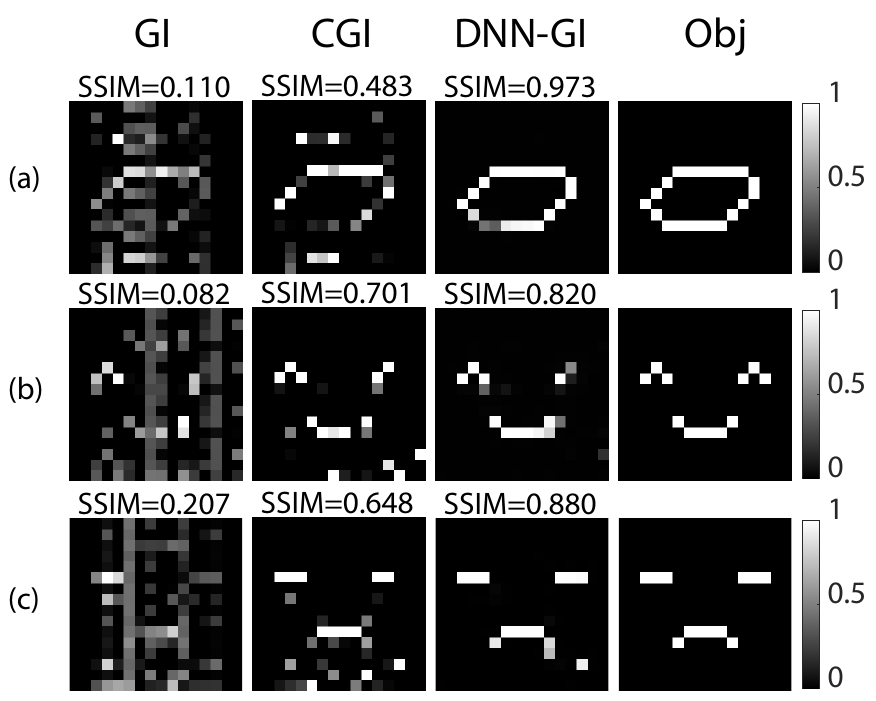}
	\caption{Neurofeedback ghost imaging results.  The four columns indicate results (from left to right) using the standard GI approach, carved GI (CGI), DNN GI reconstruction and the ground truth object.   Row (a) for the number ``0'' with 87 patterns { {(total acquisition time of 174 seconds)}}; Row (b) for a smiley face with 74 patterns { {(total acquisition time of 148 seconds)}}; and Row (c) for a sad face with 76 patterns { {(total acquisition time of 152 seconds)}}.
	}\label{ALl_imaging_results}
\end{figure}
{\bf{{ {Adaptive}} ghost imaging Results.}} Objects  are illuminated with $16\times 16$ pixel Hadamard patterns with  2 s illumination times and a flicker frequency of 6 Hz.
Hadamard carving is applied as described above followed by image reconstruction based on  standard GI, carved GI and DNN-GI.  Figure~\ref{ALl_imaging_results} shows the results obtained with the three methods for three different examples, i.e. a geometric shape and two simplified face objects (more results are shown in the SI). Image reconstruction quality is quantified by the Structural similarity index measure (SSIM), indicated above each figure. CGI, using carved patterns, performs better than traditional GI due to the carved patterns that effectively set parts of the background area to zero. \\
The best results are obtained with the DNN-GI due to the additional noise reduction that is included as part of the network structure (see SI). More importantly, whereas the straightforward GI completely fails at reconstructing an image, the DNN reconstruction applied after carving leads to high quality images with 
an average $\sim 70\%$ reduction of the total number patterns required. 
For reference, the standard GI approach (first column in Fig.~\ref{GI_result_4_4s}) required a rather prohibitive observation time for the full 256 pattern set of around $256\times2\times(16/4)^2/60=137$ minutes for the digit ``0'' in row (a), whereas the CGI and DNN-GI approaches required a total of only $87\times 2/60=3$ minutes.\\
%
\begin{figure}[t]%
	\centering
	\includegraphics[width=\columnwidth]{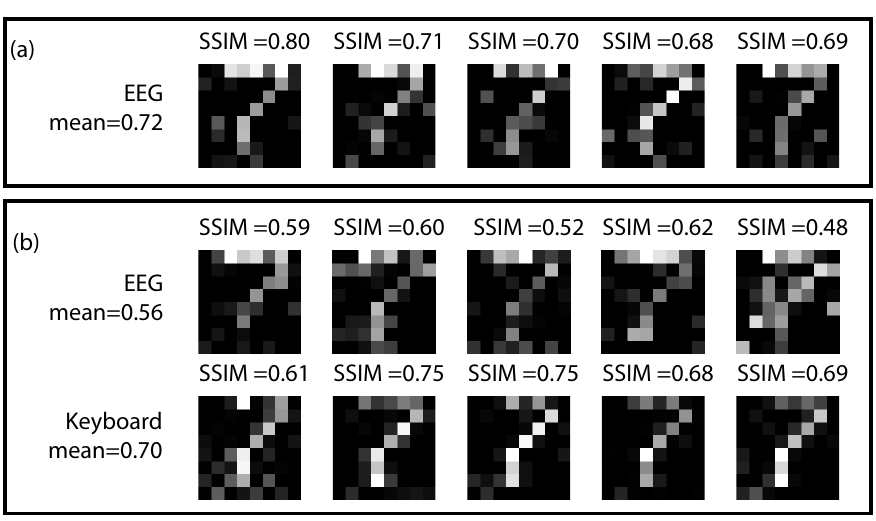}
	\caption{\label{sub}  { {Nonconscious versus conscious ghost imaging: (a) shows 5 repetitions of  `standard' (full Hadamard pattern projection) ghost imaging of the digit ``7'' using only the EEG as a read-out. The mean SSIM across the 5 repetitions is 0.72. (b) shows the case for concomitant EEG read-out and conscious read-out in the form of perceived intensity values evaluated by the participant in the range 0-15 and then typed into a keyboard (without shifting eye contact from the screen). The `conscious' processed information provides similar image quality to the EEG alone (mean SSIM = 0.70). However, the EEG reconstruction is now systematically worse and has a mean SSIM = 0.56, indicating an apparent interference between conscious processing of the data and the EEG read-out from the visual cortex.
	}} } 
\end{figure}
{\bf{Conclusions.}}
We have shown that it is possible to perform simple computational tasks that rely on brain-computer connectivity.  The chosen specific computational task can be identified as a form of ghost imaging, including an adaptive feedback approach that allows to `carve' out low-signal illumination patterns and thus significantly reduce both illumination time and final image noise.  The same approach could of course also be implemented using a standard detector. However, the emphasis here is on the possibility for the use of SSVEP recordings to establish an adaptive computational imaging scheme  where the brain is used as the sensor for image reconstruction. This lays the foundations for future work where the brain-computer system can be used for alternative forms of computation, possibly  extending also to other forms of computational imaging or also to other input channels to the brain such as the auditory system. { {As a further example of potential applications, we note that the ghost imaging protocol shown here is a form of non-explicit (or nonconscious) information processing that could be compared to an explicit processing of the information whereby the participant is asked to directly evaluate the light intensities observed on the screen. Fig.~\ref{sub} shows the results of an experiment where we compare the imaging quality quantified using a standard metric as the structure similarity index (SSIM) when processing `nonconsciously' processed data, i.e. data from the EEG, with explicit, `consciously' processed data, i.e. by asking the participant to either speak out or type the perceived light intensities in a range 0-15. Two separate `standard' ghost imaging experiments are performed in each case: (a) EEG only; (b) EEG + explicit read-out. More details are given in the SM. We systematically observed that the EEG and explicit read-out give similar (SSIM) image reconstruction qualities. However, the EEG-reconstructed image deteriorates significantly in the second set of experiments, i.e. explicitly reading out the perceived intensity values (either by verbally communicating values or typing values on a keyboard) interferes with the non-explicit read-out. This will require further study and is shown here as an example of how computational imaging tasks could be used also to gain insight into the inner workings of the detector, i.e. in this case, the human brain.}}  \\

{\bf{Acknowledgements.}}
G.W. acknowledges the support of the China Scholarship Council (CSC). D.F. acknowledges support from the Royal Academy of Engineering Chair in Emerging Technologies programme and the UK Engineering and Physical Sciences Research Council (grant EP/T00097X/). The authors would like to thank Lars Muckli for helpful discussions and suggestions.\\

{\bf{Ethical approval.}}
This research was approved by the University of Glasgow ethics approval committee, application no.300210003.\\

\bibliography{Comp_EEG}

\begin{thebibliography}{45}%
\makeatletter
\providecommand \@ifxundefined [1]{%
 \@ifx{#1\undefined}
}%
\providecommand \@ifnum [1]{%
 \ifnum #1\expandafter \@firstoftwo
 \else \expandafter \@secondoftwo
 \fi
}%
\providecommand \@ifx [1]{%
 \ifx #1\expandafter \@firstoftwo
 \else \expandafter \@secondoftwo
 \fi
}%
\providecommand \natexlab [1]{#1}%
\providecommand \enquote  [1]{``#1''}%
\providecommand \bibnamefont  [1]{#1}%
\providecommand \bibfnamefont [1]{#1}%
\providecommand \citenamefont [1]{#1}%
\providecommand \href@noop [0]{\@secondoftwo}%
\providecommand \href [0]{\begingroup \@sanitize@url \@href}%
\providecommand \@href[1]{\@@startlink{#1}\@@href}%
\providecommand \@@href[1]{\endgroup#1\@@endlink}%
\providecommand \@sanitize@url [0]{\catcode `\\12\catcode `\$12\catcode
  `\&12\catcode `\#12\catcode `\^12\catcode `\_12\catcode `\%12\relax}%
\providecommand \@@startlink[1]{}%
\providecommand \@@endlink[0]{}%
\providecommand \url  [0]{\begingroup\@sanitize@url \@url }%
\providecommand \@url [1]{\endgroup\@href {#1}{\urlprefix }}%
\providecommand \urlprefix  [0]{URL }%
\providecommand \Eprint [0]{\href }%
\providecommand \doibase [0]{http://dx.doi.org/}%
\providecommand \selectlanguage [0]{\@gobble}%
\providecommand \bibinfo  [0]{\@secondoftwo}%
\providecommand \bibfield  [0]{\@secondoftwo}%
\providecommand \translation [1]{[#1]}%
\providecommand \BibitemOpen [0]{}%
\providecommand \bibitemStop [0]{}%
\providecommand \bibitemNoStop [0]{.\EOS\space}%
\providecommand \EOS [0]{\spacefactor3000\relax}%
\providecommand \BibitemShut  [1]{\csname bibitem#1\endcsname}%
\let\auto@bib@innerbib\@empty
\bibitem [{\citenamefont {Cinel}\ \emph {et~al.}(2019)\citenamefont {Cinel},
  \citenamefont {Valeriani},\ and\ \citenamefont {Poli}}]{cinel}%
  \BibitemOpen
  \bibfield  {author} {\bibinfo {author} {\bibfnamefont {C.}~\bibnamefont
  {Cinel}}, \bibinfo {author} {\bibfnamefont {D.}~\bibnamefont {Valeriani}}, \
  and\ \bibinfo {author} {\bibfnamefont {R.}~\bibnamefont {Poli}},\ }\href
  {\doibase 10.3389/fnhum.2019.00013} {\bibfield  {journal} {\bibinfo
  {journal} {Frontiers in Human Neuroscience}\ }\textbf {\bibinfo {volume}
  {13}} (\bibinfo {year} {2019}),\ 10.3389/fnhum.2019.00013}\BibitemShut
  {NoStop}%
\bibitem [{\citenamefont {Raisamo}\ \emph {et~al.}(2019)\citenamefont
  {Raisamo}, \citenamefont {Rakkolainen}, \citenamefont {Majaranta},
  \citenamefont {Salminen}, \citenamefont {Rantala},\ and\ \citenamefont
  {Farooq}}]{raisamo}%
  \BibitemOpen
  \bibfield  {author} {\bibinfo {author} {\bibfnamefont {R.}~\bibnamefont
  {Raisamo}}, \bibinfo {author} {\bibfnamefont {I.}~\bibnamefont
  {Rakkolainen}}, \bibinfo {author} {\bibfnamefont {P.}~\bibnamefont
  {Majaranta}}, \bibinfo {author} {\bibfnamefont {K.}~\bibnamefont {Salminen}},
  \bibinfo {author} {\bibfnamefont {J.}~\bibnamefont {Rantala}}, \ and\
  \bibinfo {author} {\bibfnamefont {A.}~\bibnamefont {Farooq}},\ }\href
  {\doibase https://doi.org/10.1016/j.ijhcs.2019.05.008} {\bibfield  {journal}
  {\bibinfo  {journal} {International Journal of Human-Computer Studies}\
  }\textbf {\bibinfo {volume} {131}},\ \bibinfo {pages} {131} (\bibinfo {year}
  {2019})},\ \bibinfo {note} {50 years of the International Journal of
  Human-Computer Studies. Reflections on the past, present and future of
  human-centred technologies}\BibitemShut {NoStop}%
\bibitem [{\citenamefont {Lelievre}\ \emph {et~al.}(2013)\citenamefont
  {Lelievre}, \citenamefont {Washizawa},\ and\ \citenamefont
  {Rutkowski}}]{lelievre2013single}%
  \BibitemOpen
  \bibfield  {author} {\bibinfo {author} {\bibfnamefont {Y.}~\bibnamefont
  {Lelievre}}, \bibinfo {author} {\bibfnamefont {Y.}~\bibnamefont {Washizawa}},
  \ and\ \bibinfo {author} {\bibfnamefont {T.~M.}\ \bibnamefont {Rutkowski}},\
  }in\ \href {\doibase 10/gm9zkj} {\emph {\bibinfo {booktitle} {2013
  Asia-Pacific Signal and Information Processing Association Annual Summit and
  Conference}}}\ (\bibinfo {organization} {{IEEE}},\ \bibinfo {year} {2013})\
  pp.\ \bibinfo {pages} {1--6}\BibitemShut {NoStop}%
\bibitem [{\citenamefont {Vourvopoulos}\ and\ \citenamefont
  {Liarokapis}(2012)}]{vourvopoulos2012robot}%
  \BibitemOpen
  \bibfield  {author} {\bibinfo {author} {\bibfnamefont {A.}~\bibnamefont
  {Vourvopoulos}}\ and\ \bibinfo {author} {\bibfnamefont {F.}~\bibnamefont
  {Liarokapis}},\ }in\ \href {\doibase 10/gn34qp} {\emph {\bibinfo {booktitle}
  {2012 {{IEEE}} 11th International Conference on Trust, Security and Privacy
  in Computing and Communications}}}\ (\bibinfo {organization} {{IEEE}},\
  \bibinfo {year} {2012})\ pp.\ \bibinfo {pages} {1785--1792}\BibitemShut
  {NoStop}%
\bibitem [{\citenamefont {{van de Laar}}\ \emph {et~al.}(2013)\citenamefont
  {{van de Laar}}, \citenamefont {G{\"u}rk{\"o}k}, \citenamefont {Bos},
  \citenamefont {Poel},\ and\ \citenamefont {Nijholt}}]{van2013experiencing}%
  \BibitemOpen
  \bibfield  {author} {\bibinfo {author} {\bibfnamefont {B.}~\bibnamefont {{van
  de Laar}}}, \bibinfo {author} {\bibfnamefont {H.}~\bibnamefont
  {G{\"u}rk{\"o}k}}, \bibinfo {author} {\bibfnamefont {D.~P.-O.}\ \bibnamefont
  {Bos}}, \bibinfo {author} {\bibfnamefont {M.}~\bibnamefont {Poel}}, \ and\
  \bibinfo {author} {\bibfnamefont {A.}~\bibnamefont {Nijholt}},\ }\href
  {\doibase 10/gn34qr} {\bibfield  {journal} {\bibinfo  {journal} {IEEE
  Transactions on Computational Intelligence and AI in Games}\ }\textbf
  {\bibinfo {volume} {5}},\ \bibinfo {pages} {176} (\bibinfo {year}
  {2013})}\BibitemShut {NoStop}%
\bibitem [{\citenamefont {Bi}\ \emph {et~al.}(2013)\citenamefont {Bi},
  \citenamefont {Fan},\ and\ \citenamefont {Liu}}]{bi2013eeg}%
  \BibitemOpen
  \bibfield  {author} {\bibinfo {author} {\bibfnamefont {L.}~\bibnamefont
  {Bi}}, \bibinfo {author} {\bibfnamefont {X.-A.}\ \bibnamefont {Fan}}, \ and\
  \bibinfo {author} {\bibfnamefont {Y.}~\bibnamefont {Liu}},\ }\href {\doibase
  10/f4vhcs} {\bibfield  {journal} {\bibinfo  {journal} {IEEE transactions on
  human-machine systems}\ }\textbf {\bibinfo {volume} {43}},\ \bibinfo {pages}
  {161} (\bibinfo {year} {2013})}\BibitemShut {NoStop}%
\bibitem [{\citenamefont {Wang}\ \emph {et~al.}(2011)\citenamefont {Wang},
  \citenamefont {Degenhart}, \citenamefont {Sudre}, \citenamefont {Pomerleau},\
  and\ \citenamefont {{Tyler-Kabara}}}]{wang2011decoding}%
  \BibitemOpen
  \bibfield  {author} {\bibinfo {author} {\bibfnamefont {W.}~\bibnamefont
  {Wang}}, \bibinfo {author} {\bibfnamefont {A.~D.}\ \bibnamefont {Degenhart}},
  \bibinfo {author} {\bibfnamefont {G.~P.}\ \bibnamefont {Sudre}}, \bibinfo
  {author} {\bibfnamefont {D.~A.}\ \bibnamefont {Pomerleau}}, \ and\ \bibinfo
  {author} {\bibfnamefont {E.~C.}\ \bibnamefont {{Tyler-Kabara}}},\ }in\ \href
  {\doibase 10/d5hgn5} {\emph {\bibinfo {booktitle} {2011 Annual International
  Conference of the {{IEEE}} Engineering in Medicine and Biology Society}}}\
  (\bibinfo {organization} {{IEEE}},\ \bibinfo {year} {2011})\ pp.\ \bibinfo
  {pages} {6294--6298}\BibitemShut {NoStop}%
\bibitem [{\citenamefont {Brumberg}\ \emph {et~al.}(2010)\citenamefont
  {Brumberg}, \citenamefont {{Nieto-Castanon}}, \citenamefont {Kennedy},\ and\
  \citenamefont {Guenther}}]{brumberg2010brain}%
  \BibitemOpen
  \bibfield  {author} {\bibinfo {author} {\bibfnamefont {J.~S.}\ \bibnamefont
  {Brumberg}}, \bibinfo {author} {\bibfnamefont {A.}~\bibnamefont
  {{Nieto-Castanon}}}, \bibinfo {author} {\bibfnamefont {P.~R.}\ \bibnamefont
  {Kennedy}}, \ and\ \bibinfo {author} {\bibfnamefont {F.~H.}\ \bibnamefont
  {Guenther}},\ }\href {\doibase 10/bdgvmj} {\bibfield  {journal} {\bibinfo
  {journal} {Speech communication}\ }\textbf {\bibinfo {volume} {52}},\
  \bibinfo {pages} {367} (\bibinfo {year} {2010})}\BibitemShut {NoStop}%
\bibitem [{\citenamefont {Prataksita}\ \emph {et~al.}(2014)\citenamefont
  {Prataksita}, \citenamefont {Lin}, \citenamefont {Chou},\ and\ \citenamefont
  {Kuo}}]{prataksita2014brain}%
  \BibitemOpen
  \bibfield  {author} {\bibinfo {author} {\bibfnamefont {N.}~\bibnamefont
  {Prataksita}}, \bibinfo {author} {\bibfnamefont {Y.-T.}\ \bibnamefont {Lin}},
  \bibinfo {author} {\bibfnamefont {H.-C.}\ \bibnamefont {Chou}}, \ and\
  \bibinfo {author} {\bibfnamefont {C.-H.}\ \bibnamefont {Kuo}},\ }in\ \href
  {\doibase 10/gn34qq} {\emph {\bibinfo {booktitle} {2014 {{IEEE}}
  International Symposium on Bioelectronics and Bioinformatics ({{IEEE ISBB}}
  2014)}}}\ (\bibinfo {organization} {{IEEE}},\ \bibinfo {year} {2014})\ pp.\
  \bibinfo {pages} {1--4}\BibitemShut {NoStop}%
\bibitem [{\citenamefont {Abdulkader}\ \emph {et~al.}(2015)\citenamefont
  {Abdulkader}, \citenamefont {Atia},\ and\ \citenamefont
  {Mostafa}}]{abdulkader2015brain}%
  \BibitemOpen
  \bibfield  {author} {\bibinfo {author} {\bibfnamefont {S.~N.}\ \bibnamefont
  {Abdulkader}}, \bibinfo {author} {\bibfnamefont {A.}~\bibnamefont {Atia}}, \
  and\ \bibinfo {author} {\bibfnamefont {M.-S.~M.}\ \bibnamefont {Mostafa}},\
  }\href {\doibase 10/ggwtz3} {\bibfield  {journal} {\bibinfo  {journal}
  {Egyptian Informatics Journal}\ }\textbf {\bibinfo {volume} {16}},\ \bibinfo
  {pages} {213} (\bibinfo {year} {2015})}\BibitemShut {NoStop}%
\bibitem [{\citenamefont {Wolpaw}(2013)}]{WOLPAW201367}%
  \BibitemOpen
  \bibfield  {author} {\bibinfo {author} {\bibfnamefont {J.~R.}\ \bibnamefont
  {Wolpaw}},\ }in\ \href {\doibase 10.1016/B978-0-444-52901-5.00006-X} {\emph
  {\bibinfo {booktitle} {Neurological Rehabilitation}}},\ \bibinfo {series}
  {Handbook of Clinical Neurology}, Vol.\ \bibinfo {volume} {110},\ \bibinfo
  {editor} {edited by\ \bibinfo {editor} {\bibfnamefont {M.~P.}\ \bibnamefont
  {Barnes}}\ and\ \bibinfo {editor} {\bibfnamefont {D.~C.}\ \bibnamefont
  {Good}}}\ (\bibinfo  {publisher} {{Elsevier}},\ \bibinfo {year} {2013})\ pp.\
  \bibinfo {pages} {67--74}\BibitemShut {NoStop}%
\bibitem [{\citenamefont {Wang}\ \emph {et~al.}(2008)\citenamefont {Wang},
  \citenamefont {Gao}, \citenamefont {Hong}, \citenamefont {Jia},\ and\
  \citenamefont {Gao}}]{wang2008brain}%
  \BibitemOpen
  \bibfield  {author} {\bibinfo {author} {\bibfnamefont {Y.}~\bibnamefont
  {Wang}}, \bibinfo {author} {\bibfnamefont {X.}~\bibnamefont {Gao}}, \bibinfo
  {author} {\bibfnamefont {B.}~\bibnamefont {Hong}}, \bibinfo {author}
  {\bibfnamefont {C.}~\bibnamefont {Jia}}, \ and\ \bibinfo {author}
  {\bibfnamefont {S.}~\bibnamefont {Gao}},\ }\href {\doibase 10/d3xj2m}
  {\bibfield  {journal} {\bibinfo  {journal} {IEEE Engineering in medicine and
  biology magazine}\ }\textbf {\bibinfo {volume} {27}},\ \bibinfo {pages} {64}
  (\bibinfo {year} {2008})}\BibitemShut {NoStop}%
\bibitem [{\citenamefont {Norcia}\ \emph {et~al.}(2015)\citenamefont {Norcia},
  \citenamefont {Appelbaum}, \citenamefont {Ales}, \citenamefont {Cottereau},\
  and\ \citenamefont {Rossion}}]{norciaSteadystateVisualEvoked2015}%
  \BibitemOpen
  \bibfield  {author} {\bibinfo {author} {\bibfnamefont {A.~M.}\ \bibnamefont
  {Norcia}}, \bibinfo {author} {\bibfnamefont {L.~G.}\ \bibnamefont
  {Appelbaum}}, \bibinfo {author} {\bibfnamefont {J.~M.}\ \bibnamefont {Ales}},
  \bibinfo {author} {\bibfnamefont {B.~R.}\ \bibnamefont {Cottereau}}, \ and\
  \bibinfo {author} {\bibfnamefont {B.}~\bibnamefont {Rossion}},\ }\href
  {\doibase 10/gdz73j} {\bibfield  {journal} {\bibinfo  {journal} {Journal of
  Vision}\ }\textbf {\bibinfo {volume} {15}},\ \bibinfo {pages} {4} (\bibinfo
  {year} {2015})}\BibitemShut {NoStop}%
\bibitem [{\citenamefont {Vialatte}\ \emph {et~al.}(2010)\citenamefont
  {Vialatte}, \citenamefont {Maurice}, \citenamefont {Dauwels},\ and\
  \citenamefont {Cichocki}}]{vialatteSteadystateVisuallyEvoked2010}%
  \BibitemOpen
  \bibfield  {author} {\bibinfo {author} {\bibfnamefont {F.-B.}\ \bibnamefont
  {Vialatte}}, \bibinfo {author} {\bibfnamefont {M.}~\bibnamefont {Maurice}},
  \bibinfo {author} {\bibfnamefont {J.}~\bibnamefont {Dauwels}}, \ and\
  \bibinfo {author} {\bibfnamefont {A.}~\bibnamefont {Cichocki}},\ }\href
  {\doibase 10/fksr98} {\bibfield  {journal} {\bibinfo  {journal} {Progress in
  Neurobiology}\ }\textbf {\bibinfo {volume} {90}},\ \bibinfo {pages} {418}
  (\bibinfo {year} {2010})}\BibitemShut {NoStop}%
\bibitem [{\citenamefont {Tuncel}\ \emph {et~al.}(2019)\citenamefont {Tuncel},
  \citenamefont {Ba{\c s}aklar},\ and\ \citenamefont
  {Ider}}]{tuncelModelBasedInvestigation2019}%
  \BibitemOpen
  \bibfield  {author} {\bibinfo {author} {\bibfnamefont {Y.}~\bibnamefont
  {Tuncel}}, \bibinfo {author} {\bibfnamefont {T.}~\bibnamefont {Ba{\c
  s}aklar}}, \ and\ \bibinfo {author} {\bibfnamefont {Y.~Z.}\ \bibnamefont
  {Ider}},\ }\href {\doibase 10/gmrmx8} {\bibfield  {journal} {\bibinfo
  {journal} {Biomedical Physics \& Engineering Express}\ }\textbf {\bibinfo
  {volume} {5}},\ \bibinfo {pages} {045030} (\bibinfo {year}
  {2019})}\BibitemShut {NoStop}%
\bibitem [{\citenamefont {Kim}\ \emph {et~al.}(2011)\citenamefont {Kim},
  \citenamefont {Grabowecky}, \citenamefont {Paller},\ and\ \citenamefont
  {Suzuki}}]{kimDifferentialRolesFrequencyfollowing2011}%
  \BibitemOpen
  \bibfield  {author} {\bibinfo {author} {\bibfnamefont {Y.-J.}\ \bibnamefont
  {Kim}}, \bibinfo {author} {\bibfnamefont {M.}~\bibnamefont {Grabowecky}},
  \bibinfo {author} {\bibfnamefont {K.~A.}\ \bibnamefont {Paller}}, \ and\
  \bibinfo {author} {\bibfnamefont {S.}~\bibnamefont {Suzuki}},\ }\href
  {\doibase 10/c57kmp} {\bibfield  {journal} {\bibinfo  {journal} {Journal of
  Cognitive Neuroscience}\ }\textbf {\bibinfo {volume} {23}},\ \bibinfo {pages}
  {1875} (\bibinfo {year} {2011})}\BibitemShut {NoStop}%
\bibitem [{\citenamefont {Labecki}\ \emph {et~al.}(2016)\citenamefont
  {Labecki}, \citenamefont {Kus}, \citenamefont {Brzozowska}, \citenamefont
  {Stacewicz}, \citenamefont {Bhattacharya},\ and\ \citenamefont
  {Suffczynski}}]{labeckiNonlinearOriginSSVEP2016}%
  \BibitemOpen
  \bibfield  {author} {\bibinfo {author} {\bibfnamefont {M.}~\bibnamefont
  {Labecki}}, \bibinfo {author} {\bibfnamefont {R.}~\bibnamefont {Kus}},
  \bibinfo {author} {\bibfnamefont {A.}~\bibnamefont {Brzozowska}}, \bibinfo
  {author} {\bibfnamefont {T.}~\bibnamefont {Stacewicz}}, \bibinfo {author}
  {\bibfnamefont {B.~S.}\ \bibnamefont {Bhattacharya}}, \ and\ \bibinfo
  {author} {\bibfnamefont {P.}~\bibnamefont {Suffczynski}},\ }\href {\doibase
  10/gdz72w} {\bibfield  {journal} {\bibinfo  {journal} {Frontiers in
  Computational Neuroscience}\ }\textbf {\bibinfo {volume} {10}},\ \bibinfo
  {pages} {129} (\bibinfo {year} {2016})}\BibitemShut {NoStop}%
\bibitem [{\citenamefont {Zhao}\ \emph {et~al.}(2012)\citenamefont {Zhao},
  \citenamefont {Gong}, \citenamefont {Chen}, \citenamefont {Li}, \citenamefont
  {Wang}, \citenamefont {Xu},\ and\ \citenamefont
  {Han}}]{zhaoGhostImagingLidar2012}%
  \BibitemOpen
  \bibfield  {author} {\bibinfo {author} {\bibfnamefont {C.~Q.}\ \bibnamefont
  {Zhao}}, \bibinfo {author} {\bibfnamefont {W.~L.}\ \bibnamefont {Gong}},
  \bibinfo {author} {\bibfnamefont {M.~L.}\ \bibnamefont {Chen}}, \bibinfo
  {author} {\bibfnamefont {E.~R.}\ \bibnamefont {Li}}, \bibinfo {author}
  {\bibfnamefont {H.}~\bibnamefont {Wang}}, \bibinfo {author} {\bibfnamefont
  {W.~D.}\ \bibnamefont {Xu}}, \ and\ \bibinfo {author} {\bibfnamefont {S.~S.}\
  \bibnamefont {Han}},\ }\href {\doibase 10.1063/1.4757874} {\bibfield
  {journal} {\bibinfo  {journal} {Applied Physics Letters}\ }\textbf {\bibinfo
  {volume} {101}},\ \bibinfo {pages} {141123} (\bibinfo {year}
  {2012})}\BibitemShut {NoStop}%
\bibitem [{\citenamefont {Pittman}\ \emph {et~al.}(1995)\citenamefont
  {Pittman}, \citenamefont {Shih}, \citenamefont {Strekalov},\ and\
  \citenamefont {Sergienko}}]{pittmanOpticalImagingMeans1995}%
  \BibitemOpen
  \bibfield  {author} {\bibinfo {author} {\bibfnamefont {T.~B.}\ \bibnamefont
  {Pittman}}, \bibinfo {author} {\bibfnamefont {Y.~H.}\ \bibnamefont {Shih}},
  \bibinfo {author} {\bibfnamefont {D.~V.}\ \bibnamefont {Strekalov}}, \ and\
  \bibinfo {author} {\bibfnamefont {A.~V.}\ \bibnamefont {Sergienko}},\ }\href
  {\doibase 10/bpcfqx} {\bibfield  {journal} {\bibinfo  {journal} {Physical
  Review A}\ }\textbf {\bibinfo {volume} {52}},\ \bibinfo {pages} {R3429}
  (\bibinfo {year} {1995})}\BibitemShut {NoStop}%
\bibitem [{\citenamefont {Zhang}\ \emph {et~al.}(2005)\citenamefont {Zhang},
  \citenamefont {Zhai}, \citenamefont {Wu},\ and\ \citenamefont
  {Chen}}]{zhangCorrelatedTwophotonImaging2005}%
  \BibitemOpen
  \bibfield  {author} {\bibinfo {author} {\bibfnamefont {D.}~\bibnamefont
  {Zhang}}, \bibinfo {author} {\bibfnamefont {Y.~H.}\ \bibnamefont {Zhai}},
  \bibinfo {author} {\bibfnamefont {L.~A.}\ \bibnamefont {Wu}}, \ and\ \bibinfo
  {author} {\bibfnamefont {X.~H.}\ \bibnamefont {Chen}},\ }\href {\doibase
  10/dbsd3p} {\bibfield  {journal} {\bibinfo  {journal} {Optics Letters}\
  }\textbf {\bibinfo {volume} {30}},\ \bibinfo {pages} {2354} (\bibinfo {year}
  {2005})}\BibitemShut {NoStop}%
\bibitem [{\citenamefont {Pelliccia}\ \emph {et~al.}(2016)\citenamefont
  {Pelliccia}, \citenamefont {Rack}, \citenamefont {Scheel}, \citenamefont
  {Cantelli},\ and\ \citenamefont
  {Paganin}}]{pellicciaExperimentalXrayGhost2016}%
  \BibitemOpen
  \bibfield  {author} {\bibinfo {author} {\bibfnamefont {D.}~\bibnamefont
  {Pelliccia}}, \bibinfo {author} {\bibfnamefont {A.}~\bibnamefont {Rack}},
  \bibinfo {author} {\bibfnamefont {M.}~\bibnamefont {Scheel}}, \bibinfo
  {author} {\bibfnamefont {V.}~\bibnamefont {Cantelli}}, \ and\ \bibinfo
  {author} {\bibfnamefont {D.~M.}\ \bibnamefont {Paganin}},\ }\href {\doibase
  10/gfgxj5} {\bibfield  {journal} {\bibinfo  {journal} {Physical review
  letters}\ }\textbf {\bibinfo {volume} {117}},\ \bibinfo {pages} {113902}
  (\bibinfo {year} {2016})}\BibitemShut {NoStop}%
\bibitem [{\citenamefont {Xu}\ \emph {et~al.}(2018)\citenamefont {Xu},
  \citenamefont {Chen}, \citenamefont {Penuelas}, \citenamefont {Padgett},\
  and\ \citenamefont {Sun}}]{xu1000FpsComputational2018}%
  \BibitemOpen
  \bibfield  {author} {\bibinfo {author} {\bibfnamefont {Z.~H.}\ \bibnamefont
  {Xu}}, \bibinfo {author} {\bibfnamefont {W.}~\bibnamefont {Chen}}, \bibinfo
  {author} {\bibfnamefont {J.}~\bibnamefont {Penuelas}}, \bibinfo {author}
  {\bibfnamefont {M.}~\bibnamefont {Padgett}}, \ and\ \bibinfo {author}
  {\bibfnamefont {M.~J.}\ \bibnamefont {Sun}},\ }\href {\doibase 10/gf4c8s}
  {\bibfield  {journal} {\bibinfo  {journal} {Optics Express}\ }\textbf
  {\bibinfo {volume} {26}},\ \bibinfo {pages} {2427} (\bibinfo {year}
  {2018})}\BibitemShut {NoStop}%
\bibitem [{\citenamefont {Zhao}\ \emph {et~al.}(2019)\citenamefont {Zhao},
  \citenamefont {Chen}, \citenamefont {Yuan}, \citenamefont {Zheng},
  \citenamefont {Liu}, \citenamefont {Xu},\ and\ \citenamefont
  {Zhou}}]{zhao2019ultrahigh}%
  \BibitemOpen
  \bibfield  {author} {\bibinfo {author} {\bibfnamefont {W.}~\bibnamefont
  {Zhao}}, \bibinfo {author} {\bibfnamefont {H.}~\bibnamefont {Chen}}, \bibinfo
  {author} {\bibfnamefont {Y.}~\bibnamefont {Yuan}}, \bibinfo {author}
  {\bibfnamefont {H.}~\bibnamefont {Zheng}}, \bibinfo {author} {\bibfnamefont
  {J.}~\bibnamefont {Liu}}, \bibinfo {author} {\bibfnamefont {Z.}~\bibnamefont
  {Xu}}, \ and\ \bibinfo {author} {\bibfnamefont {Y.}~\bibnamefont {Zhou}},\
  }\href {\doibase 10/gn67dv} {\bibfield  {journal} {\bibinfo  {journal}
  {Physical Review Applied}\ }\textbf {\bibinfo {volume} {12}},\ \bibinfo
  {pages} {034049} (\bibinfo {year} {2019})}\BibitemShut {NoStop}%
\bibitem [{\citenamefont {Katkovnik}\ and\ \citenamefont
  {Astola}(2012)}]{katkovnikCompressiveSensingComputational2012}%
  \BibitemOpen
  \bibfield  {author} {\bibinfo {author} {\bibfnamefont {V.}~\bibnamefont
  {Katkovnik}}\ and\ \bibinfo {author} {\bibfnamefont {J.}~\bibnamefont
  {Astola}},\ }\href {\doibase 10/gf7pct} {\bibfield  {journal} {\bibinfo
  {journal} {JOSA A}\ }\textbf {\bibinfo {volume} {29}},\ \bibinfo {pages}
  {1556} (\bibinfo {year} {2012})}\BibitemShut {NoStop}%
\bibitem [{\citenamefont {Sun}\ \emph {et~al.}(2012)\citenamefont {Sun},
  \citenamefont {Welsh}, \citenamefont {Edgar}, \citenamefont {Shapiro},\ and\
  \citenamefont {Padgett}}]{sun2012normalized}%
  \BibitemOpen
  \bibfield  {author} {\bibinfo {author} {\bibfnamefont {B.}~\bibnamefont
  {Sun}}, \bibinfo {author} {\bibfnamefont {S.~S.}\ \bibnamefont {Welsh}},
  \bibinfo {author} {\bibfnamefont {M.~P.}\ \bibnamefont {Edgar}}, \bibinfo
  {author} {\bibfnamefont {J.~H.}\ \bibnamefont {Shapiro}}, \ and\ \bibinfo
  {author} {\bibfnamefont {M.~J.}\ \bibnamefont {Padgett}},\ }\href {\doibase
  10/gf7pcv} {\bibfield  {journal} {\bibinfo  {journal} {Optics Express}\
  }\textbf {\bibinfo {volume} {20}},\ \bibinfo {pages} {16892} (\bibinfo {year}
  {2012})}\BibitemShut {NoStop}%
\bibitem [{\citenamefont {Yao}\ \emph {et~al.}(2014)\citenamefont {Yao},
  \citenamefont {Yu}, \citenamefont {Liu}, \citenamefont {Li}, \citenamefont
  {Li}, \citenamefont {Wu},\ and\ \citenamefont
  {Zhai}}]{yaoIterativeDenoisingGhost2014}%
  \BibitemOpen
  \bibfield  {author} {\bibinfo {author} {\bibfnamefont {X.~R.}\ \bibnamefont
  {Yao}}, \bibinfo {author} {\bibfnamefont {W.~K.}\ \bibnamefont {Yu}},
  \bibinfo {author} {\bibfnamefont {X.~F.}\ \bibnamefont {Liu}}, \bibinfo
  {author} {\bibfnamefont {L.~Z.}\ \bibnamefont {Li}}, \bibinfo {author}
  {\bibfnamefont {M.~F.}\ \bibnamefont {Li}}, \bibinfo {author} {\bibfnamefont
  {L.~A.}\ \bibnamefont {Wu}}, \ and\ \bibinfo {author} {\bibfnamefont {G.~J.}\
  \bibnamefont {Zhai}},\ }\href {\doibase 10.1364/oe.22.024268} {\bibfield
  {journal} {\bibinfo  {journal} {Optics Express}\ }\textbf {\bibinfo {volume}
  {22}},\ \bibinfo {pages} {24268} (\bibinfo {year} {2014})}\BibitemShut
  {NoStop}%
\bibitem [{\citenamefont {He}\ \emph {et~al.}(2018)\citenamefont {He},
  \citenamefont {Wang}, \citenamefont {Dong}, \citenamefont {Zhu},
  \citenamefont {Chen}, \citenamefont {Zhang},\ and\ \citenamefont
  {Xu}}]{heGhostImagingBased2018}%
  \BibitemOpen
  \bibfield  {author} {\bibinfo {author} {\bibfnamefont {Y.}~\bibnamefont
  {He}}, \bibinfo {author} {\bibfnamefont {G.}~\bibnamefont {Wang}}, \bibinfo
  {author} {\bibfnamefont {G.}~\bibnamefont {Dong}}, \bibinfo {author}
  {\bibfnamefont {S.}~\bibnamefont {Zhu}}, \bibinfo {author} {\bibfnamefont
  {H.}~\bibnamefont {Chen}}, \bibinfo {author} {\bibfnamefont {A.}~\bibnamefont
  {Zhang}}, \ and\ \bibinfo {author} {\bibfnamefont {Z.}~\bibnamefont {Xu}},\
  }\href {\doibase 10.1038/s41598-018-24731-2} {\bibfield  {journal} {\bibinfo
  {journal} {Sci Rep}\ }\textbf {\bibinfo {volume} {8}},\ \bibinfo {pages}
  {6469} (\bibinfo {year} {2018})}\BibitemShut {NoStop}%
\bibitem [{\citenamefont {Shimobaba}\ \emph {et~al.}(2018)\citenamefont
  {Shimobaba}, \citenamefont {Endo}, \citenamefont {Nishitsuji}, \citenamefont
  {Takahashi}, \citenamefont {Nagahama}, \citenamefont {Hasegawa},
  \citenamefont {Sano}, \citenamefont {Hirayama}, \citenamefont {Kakue},
  \citenamefont {Shiraki},\ and\ \citenamefont
  {Ito}}]{shimobabaComputationalGhostImaging2018}%
  \BibitemOpen
  \bibfield  {author} {\bibinfo {author} {\bibfnamefont {T.}~\bibnamefont
  {Shimobaba}}, \bibinfo {author} {\bibfnamefont {Y.}~\bibnamefont {Endo}},
  \bibinfo {author} {\bibfnamefont {T.}~\bibnamefont {Nishitsuji}}, \bibinfo
  {author} {\bibfnamefont {T.}~\bibnamefont {Takahashi}}, \bibinfo {author}
  {\bibfnamefont {Y.}~\bibnamefont {Nagahama}}, \bibinfo {author}
  {\bibfnamefont {S.}~\bibnamefont {Hasegawa}}, \bibinfo {author}
  {\bibfnamefont {M.}~\bibnamefont {Sano}}, \bibinfo {author} {\bibfnamefont
  {R.}~\bibnamefont {Hirayama}}, \bibinfo {author} {\bibfnamefont
  {T.}~\bibnamefont {Kakue}}, \bibinfo {author} {\bibfnamefont
  {A.}~\bibnamefont {Shiraki}}, \ and\ \bibinfo {author} {\bibfnamefont
  {T.}~\bibnamefont {Ito}},\ }\href {\doibase 10/gf7pdb} {\bibfield  {journal}
  {\bibinfo  {journal} {Optics Communications}\ }\textbf {\bibinfo {volume}
  {413}},\ \bibinfo {pages} {147} (\bibinfo {year} {2018})}\BibitemShut
  {NoStop}%
\bibitem [{\citenamefont {Moreau}\ \emph {et~al.}(2018)\citenamefont {Moreau},
  \citenamefont {Toninelli}, \citenamefont {Gregory},\ and\ \citenamefont
  {Padgett}}]{moreauGhostImagingUsing2018a}%
  \BibitemOpen
  \bibfield  {author} {\bibinfo {author} {\bibfnamefont {P.-A.}\ \bibnamefont
  {Moreau}}, \bibinfo {author} {\bibfnamefont {E.}~\bibnamefont {Toninelli}},
  \bibinfo {author} {\bibfnamefont {T.}~\bibnamefont {Gregory}}, \ and\
  \bibinfo {author} {\bibfnamefont {M.~J.}\ \bibnamefont {Padgett}},\ }\href
  {\doibase 10/ggbrtb} {\bibfield  {journal} {\bibinfo  {journal} {Laser \&
  Photonics Reviews}\ }\textbf {\bibinfo {volume} {12}},\ \bibinfo {pages}
  {1700143} (\bibinfo {year} {2018})}\BibitemShut {NoStop}%
\bibitem [{\citenamefont {Padgett}\ and\ \citenamefont
  {Boyd}(2017)}]{padgett2017introduction}%
  \BibitemOpen
  \bibfield  {author} {\bibinfo {author} {\bibfnamefont {M.~J.}\ \bibnamefont
  {Padgett}}\ and\ \bibinfo {author} {\bibfnamefont {R.~W.}\ \bibnamefont
  {Boyd}},\ }\href {\doibase 10/gf9sp5} {\bibfield  {journal} {\bibinfo
  {journal} {Philosophical Transactions of the Royal Society A: Mathematical,
  Physical and Engineering Sciences}\ }\textbf {\bibinfo {volume} {375}},\
  \bibinfo {pages} {20160233} (\bibinfo {year} {2017})}\BibitemShut {NoStop}%
\bibitem [{\citenamefont {Padgett}\ \emph {et~al.}(2016)\citenamefont
  {Padgett}, \citenamefont {Aspden}, \citenamefont {Gibson}, \citenamefont
  {Edgar},\ and\ \citenamefont {Spalding}}]{padgett2016ghost}%
  \BibitemOpen
  \bibfield  {author} {\bibinfo {author} {\bibfnamefont {M.}~\bibnamefont
  {Padgett}}, \bibinfo {author} {\bibfnamefont {R.}~\bibnamefont {Aspden}},
  \bibinfo {author} {\bibfnamefont {G.}~\bibnamefont {Gibson}}, \bibinfo
  {author} {\bibfnamefont {M.}~\bibnamefont {Edgar}}, \ and\ \bibinfo {author}
  {\bibfnamefont {G.}~\bibnamefont {Spalding}},\ }\href {\doibase 10/gn478b}
  {\bibfield  {journal} {\bibinfo  {journal} {Optics and Photonics News}\
  }\textbf {\bibinfo {volume} {27}},\ \bibinfo {pages} {38} (\bibinfo {year}
  {2016})}\BibitemShut {NoStop}%
\bibitem [{\citenamefont
  {Shapiro}(2008)}]{shapiroComputationalGhostImaging2008}%
  \BibitemOpen
  \bibfield  {author} {\bibinfo {author} {\bibfnamefont {J.~H.}\ \bibnamefont
  {Shapiro}},\ }\href {\doibase 10/dg9qks} {\bibfield  {journal} {\bibinfo
  {journal} {Physical Review A}\ }\textbf {\bibinfo {volume} {78}},\ \bibinfo
  {pages} {061802} (\bibinfo {year} {2008})}\BibitemShut {NoStop}%
\bibitem [{\citenamefont {Bai}\ \emph {et~al.}(2017)\citenamefont {Bai},
  \citenamefont {He}, \citenamefont {Liu}, \citenamefont {Zhou}, \citenamefont
  {Zheng}, \citenamefont {Zhang},\ and\ \citenamefont
  {Xu}}]{baiImagingCornersSinglepixel2017}%
  \BibitemOpen
  \bibfield  {author} {\bibinfo {author} {\bibfnamefont {B.}~\bibnamefont
  {Bai}}, \bibinfo {author} {\bibfnamefont {Y.}~\bibnamefont {He}}, \bibinfo
  {author} {\bibfnamefont {J.}~\bibnamefont {Liu}}, \bibinfo {author}
  {\bibfnamefont {Y.}~\bibnamefont {Zhou}}, \bibinfo {author} {\bibfnamefont
  {H.}~\bibnamefont {Zheng}}, \bibinfo {author} {\bibfnamefont
  {S.}~\bibnamefont {Zhang}}, \ and\ \bibinfo {author} {\bibfnamefont
  {Z.}~\bibnamefont {Xu}},\ }\href {\doibase 10/gb4rfm} {\bibfield  {journal}
  {\bibinfo  {journal} {Optik}\ }\textbf {\bibinfo {volume} {147}},\ \bibinfo
  {pages} {136} (\bibinfo {year} {2017})}\BibitemShut {NoStop}%
\bibitem [{\citenamefont {Faccio}\ \emph {et~al.}(2020)\citenamefont {Faccio},
  \citenamefont {Velten},\ and\ \citenamefont
  {Wetzstein}}]{faccioNonlineofsightImaging2020}%
  \BibitemOpen
  \bibfield  {author} {\bibinfo {author} {\bibfnamefont {D.}~\bibnamefont
  {Faccio}}, \bibinfo {author} {\bibfnamefont {A.}~\bibnamefont {Velten}}, \
  and\ \bibinfo {author} {\bibfnamefont {G.}~\bibnamefont {Wetzstein}},\ }\href
  {\doibase 10/ggznwb} {\bibfield  {journal} {\bibinfo  {journal} {Nature
  Reviews Physics}\ }\textbf {\bibinfo {volume} {2}},\ \bibinfo {pages} {318}
  (\bibinfo {year} {2020})}\BibitemShut {NoStop}%
\bibitem [{\citenamefont {Saunders}\ \emph {et~al.}(2019)\citenamefont
  {Saunders}, \citenamefont {{Murray-Bruce}},\ and\ \citenamefont
  {Goyal}}]{saundersComputationalPeriscopyOrdinary2019}%
  \BibitemOpen
  \bibfield  {author} {\bibinfo {author} {\bibfnamefont {C.}~\bibnamefont
  {Saunders}}, \bibinfo {author} {\bibfnamefont {J.}~\bibnamefont
  {{Murray-Bruce}}}, \ and\ \bibinfo {author} {\bibfnamefont {V.~K.}\
  \bibnamefont {Goyal}},\ }\href {\doibase 10/gftrpd} {\bibfield  {journal}
  {\bibinfo  {journal} {Nature}\ }\textbf {\bibinfo {volume} {565}},\ \bibinfo
  {pages} {472} (\bibinfo {year} {2019})}\BibitemShut {NoStop}%
\bibitem [{\citenamefont {O'Toole}\ \emph {et~al.}(2018)\citenamefont
  {O'Toole}, \citenamefont {Lindell},\ and\ \citenamefont
  {Wetzstein}}]{otooleConfocalNonlineofsightImaging2018}%
  \BibitemOpen
  \bibfield  {author} {\bibinfo {author} {\bibfnamefont {M.}~\bibnamefont
  {O'Toole}}, \bibinfo {author} {\bibfnamefont {D.~B.}\ \bibnamefont
  {Lindell}}, \ and\ \bibinfo {author} {\bibfnamefont {G.}~\bibnamefont
  {Wetzstein}},\ }\href {\doibase 10/gc3gs5} {\bibfield  {journal} {\bibinfo
  {journal} {Nature}\ }\textbf {\bibinfo {volume} {555}},\ \bibinfo {pages}
  {338} (\bibinfo {year} {2018})}\BibitemShut {NoStop}%
\bibitem [{\citenamefont {Velten}\ \emph {et~al.}(2012)\citenamefont {Velten},
  \citenamefont {Willwacher}, \citenamefont {Gupta}, \citenamefont
  {Veeraraghavan}, \citenamefont {Bawendi},\ and\ \citenamefont
  {Raskar}}]{veltenRecoveringThreedimensionalShape2012}%
  \BibitemOpen
  \bibfield  {author} {\bibinfo {author} {\bibfnamefont {A.}~\bibnamefont
  {Velten}}, \bibinfo {author} {\bibfnamefont {T.}~\bibnamefont {Willwacher}},
  \bibinfo {author} {\bibfnamefont {O.}~\bibnamefont {Gupta}}, \bibinfo
  {author} {\bibfnamefont {A.}~\bibnamefont {Veeraraghavan}}, \bibinfo {author}
  {\bibfnamefont {M.~G.}\ \bibnamefont {Bawendi}}, \ and\ \bibinfo {author}
  {\bibfnamefont {R.}~\bibnamefont {Raskar}},\ }\href {\doibase 10/ggz3qh}
  {\bibfield  {journal} {\bibinfo  {journal} {Nature Communications}\ }\textbf
  {\bibinfo {volume} {3}},\ \bibinfo {pages} {745} (\bibinfo {year}
  {2012})}\BibitemShut {NoStop}%
\bibitem [{\citenamefont {Chen}\ \emph {et~al.}(2020)\citenamefont {Chen},
  \citenamefont {Wei}, \citenamefont {Kutulakos}, \citenamefont
  {Rusinkiewicz},\ and\ \citenamefont
  {Heide}}]{chenLearnedFeatureEmbeddings2020}%
  \BibitemOpen
  \bibfield  {author} {\bibinfo {author} {\bibfnamefont {W.}~\bibnamefont
  {Chen}}, \bibinfo {author} {\bibfnamefont {F.}~\bibnamefont {Wei}}, \bibinfo
  {author} {\bibfnamefont {K.~N.}\ \bibnamefont {Kutulakos}}, \bibinfo {author}
  {\bibfnamefont {S.}~\bibnamefont {Rusinkiewicz}}, \ and\ \bibinfo {author}
  {\bibfnamefont {F.}~\bibnamefont {Heide}},\ }\href {\doibase 10/ghsr98}
  {\bibfield  {journal} {\bibinfo  {journal} {ACM Transactions on Graphics}\
  }\textbf {\bibinfo {volume} {39}},\ \bibinfo {pages} {1} (\bibinfo {year}
  {2020})}\BibitemShut {NoStop}%
\bibitem [{\citenamefont {Boccolini}\ \emph {et~al.}(2019)\citenamefont
  {Boccolini}, \citenamefont {Fedrizzi},\ and\ \citenamefont
  {Faccio}}]{boccoliniGhostImagingHuman2019}%
  \BibitemOpen
  \bibfield  {author} {\bibinfo {author} {\bibfnamefont {A.}~\bibnamefont
  {Boccolini}}, \bibinfo {author} {\bibfnamefont {A.}~\bibnamefont {Fedrizzi}},
  \ and\ \bibinfo {author} {\bibfnamefont {D.}~\bibnamefont {Faccio}},\ }\href
  {\doibase 10/gf7pc4} {\bibfield  {journal} {\bibinfo  {journal} {Optics
  Express}\ }\textbf {\bibinfo {volume} {27}},\ \bibinfo {pages} {9258}
  (\bibinfo {year} {2019})}\BibitemShut {NoStop}%
\bibitem [{\citenamefont {Wang}\ \emph
  {et~al.}(2020{\natexlab{a}})\citenamefont {Wang}, \citenamefont {Zheng},
  \citenamefont {Tang}, \citenamefont {Zhou}, \citenamefont {Chen},
  \citenamefont {Liu}, \citenamefont {He}, \citenamefont {Yuan}, \citenamefont
  {Li},\ and\ \citenamefont {Xu}}]{wangAllOpticalNakedEyeGhost2020}%
  \BibitemOpen
  \bibfield  {author} {\bibinfo {author} {\bibfnamefont {G.}~\bibnamefont
  {Wang}}, \bibinfo {author} {\bibfnamefont {H.}~\bibnamefont {Zheng}},
  \bibinfo {author} {\bibfnamefont {Z.}~\bibnamefont {Tang}}, \bibinfo {author}
  {\bibfnamefont {Y.}~\bibnamefont {Zhou}}, \bibinfo {author} {\bibfnamefont
  {H.}~\bibnamefont {Chen}}, \bibinfo {author} {\bibfnamefont {J.}~\bibnamefont
  {Liu}}, \bibinfo {author} {\bibfnamefont {Y.}~\bibnamefont {He}}, \bibinfo
  {author} {\bibfnamefont {Y.}~\bibnamefont {Yuan}}, \bibinfo {author}
  {\bibfnamefont {F.}~\bibnamefont {Li}}, \ and\ \bibinfo {author}
  {\bibfnamefont {Z.}~\bibnamefont {Xu}},\ }\href {\doibase 10/ggn9qv}
  {\bibfield  {journal} {\bibinfo  {journal} {Scientific Reports}\ }\textbf
  {\bibinfo {volume} {10}},\ \bibinfo {pages} {2493} (\bibinfo {year}
  {2020}{\natexlab{a}})}\BibitemShut {NoStop}%
\bibitem [{\citenamefont {Wang}\ \emph
  {et~al.}(2020{\natexlab{b}})\citenamefont {Wang}, \citenamefont {Zheng},
  \citenamefont {Tang}, \citenamefont {He}, \citenamefont {Zhou}, \citenamefont
  {Chen}, \citenamefont {Liu}, \citenamefont {Yuan}, \citenamefont {Li},\ and\
  \citenamefont {Xu}}]{wangNakedeyeGhostImaging2020}%
  \BibitemOpen
  \bibfield  {author} {\bibinfo {author} {\bibfnamefont {G.}~\bibnamefont
  {Wang}}, \bibinfo {author} {\bibfnamefont {H.}~\bibnamefont {Zheng}},
  \bibinfo {author} {\bibfnamefont {Z.}~\bibnamefont {Tang}}, \bibinfo {author}
  {\bibfnamefont {Y.}~\bibnamefont {He}}, \bibinfo {author} {\bibfnamefont
  {Y.}~\bibnamefont {Zhou}}, \bibinfo {author} {\bibfnamefont {H.}~\bibnamefont
  {Chen}}, \bibinfo {author} {\bibfnamefont {J.}~\bibnamefont {Liu}}, \bibinfo
  {author} {\bibfnamefont {Y.}~\bibnamefont {Yuan}}, \bibinfo {author}
  {\bibfnamefont {F.}~\bibnamefont {Li}}, \ and\ \bibinfo {author}
  {\bibfnamefont {Z.}~\bibnamefont {Xu}},\ }\href {\doibase 10/gjnwbt}
  {\bibfield  {journal} {\bibinfo  {journal} {Chinese Optics Letters}\ }\textbf
  {\bibinfo {volume} {18}},\ \bibinfo {pages} {091101} (\bibinfo {year}
  {2020}{\natexlab{b}})}\BibitemShut {NoStop}%
\bibitem [{\citenamefont {Auger}\ and\ \citenamefont
  {Flandrin}(1995)}]{auger1995improving}%
  \BibitemOpen
  \bibfield  {author} {\bibinfo {author} {\bibfnamefont {F.}~\bibnamefont
  {Auger}}\ and\ \bibinfo {author} {\bibfnamefont {P.}~\bibnamefont
  {Flandrin}},\ }\href {\doibase 10/dpmrhp} {\bibfield  {journal} {\bibinfo
  {journal} {IEEE Transactions on signal processing}\ }\textbf {\bibinfo
  {volume} {43}},\ \bibinfo {pages} {1068} (\bibinfo {year}
  {1995})}\BibitemShut {NoStop}%
\bibitem [{\citenamefont {Fulop}\ and\ \citenamefont
  {Fitz}(2006)}]{fulop2006algorithms}%
  \BibitemOpen
  \bibfield  {author} {\bibinfo {author} {\bibfnamefont {S.~A.}\ \bibnamefont
  {Fulop}}\ and\ \bibinfo {author} {\bibfnamefont {K.}~\bibnamefont {Fitz}},\
  }\href {\doibase 10/bghmxx} {\bibfield  {journal} {\bibinfo  {journal} {The
  Journal of the Acoustical Society of America}\ }\textbf {\bibinfo {volume}
  {119}},\ \bibinfo {pages} {360} (\bibinfo {year} {2006})}\BibitemShut
  {NoStop}%
\bibitem [{\citenamefont {Zhang}\ \emph {et~al.}(2017)\citenamefont {Zhang},
  \citenamefont {Zuo}, \citenamefont {Chen}, \citenamefont {Meng},\ and\
  \citenamefont {Zhang}}]{zhang2017beyond}%
  \BibitemOpen
  \bibfield  {author} {\bibinfo {author} {\bibfnamefont {K.}~\bibnamefont
  {Zhang}}, \bibinfo {author} {\bibfnamefont {W.}~\bibnamefont {Zuo}}, \bibinfo
  {author} {\bibfnamefont {Y.}~\bibnamefont {Chen}}, \bibinfo {author}
  {\bibfnamefont {D.}~\bibnamefont {Meng}}, \ and\ \bibinfo {author}
  {\bibfnamefont {L.}~\bibnamefont {Zhang}},\ }\href {\doibase 10/f9864f}
  {\bibfield  {journal} {\bibinfo  {journal} {IEEE transactions on image
  processing}\ }\textbf {\bibinfo {volume} {26}},\ \bibinfo {pages} {3142}
  (\bibinfo {year} {2017})}\BibitemShut {NoStop}%
\bibitem [{\citenamefont {Johnson}\ \emph {et~al.}(2020)\citenamefont
  {Johnson}, \citenamefont {Moreau}, \citenamefont {Gregory},\ and\
  \citenamefont {Padgett}}]{padgett2020}%
  \BibitemOpen
  \bibfield  {author} {\bibinfo {author} {\bibfnamefont {S.~D.}\ \bibnamefont
  {Johnson}}, \bibinfo {author} {\bibfnamefont {P.-A.}\ \bibnamefont {Moreau}},
  \bibinfo {author} {\bibfnamefont {T.}~\bibnamefont {Gregory}}, \ and\
  \bibinfo {author} {\bibfnamefont {M.~J.}\ \bibnamefont {Padgett}},\ }\href
  {\doibase 10/gg6shm} {\bibfield  {journal} {\bibinfo  {journal} {Applied
  Physics Letters}\ }\textbf {\bibinfo {volume} {116}},\ \bibinfo {pages}
  {260504} (\bibinfo {year} {2020})},\ \Eprint
  {http://arxiv.org/abs/https://doi.org/10.1063/5.0009493}
  {https://doi.org/10.1063/5.0009493} \BibitemShut {NoStop}%
\end{thebibliography}%

\clearpage

\section*{Supplementary information.}
\begin{figure}[ht]%
\centering
\includegraphics[width=\columnwidth]{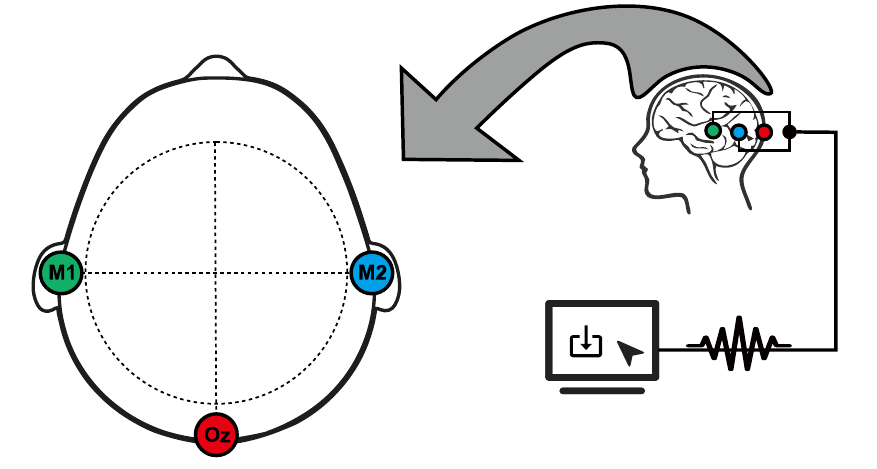}
\caption{The layout of EEG recording system.}\label{EEG_sys}
\end{figure}

\begin{figure}[ht]%
\centering
\includegraphics[width=\columnwidth]{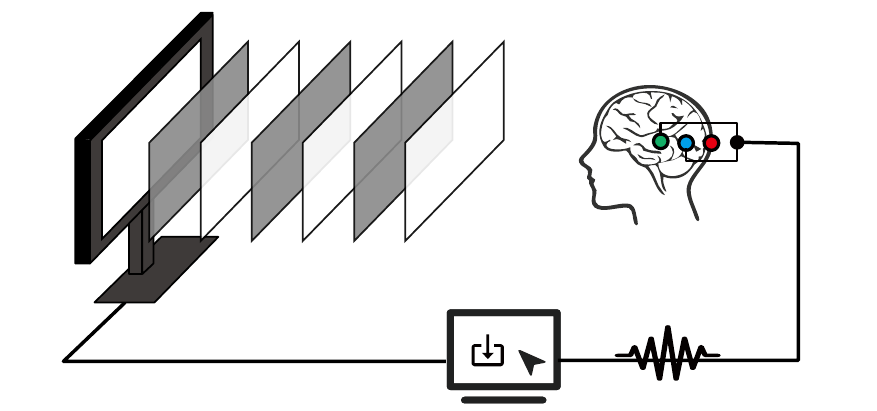}
\caption{EEG system with LCD screen stimulus.} \label{Screen_EEG_sys}
\end{figure}

{\bf{EEG setup.}}
A schematic overview is shown in Fig.~\ref{EEG_sys}: we use a 3 pole EEG device, which only has one active electrode located in Oz  (medial occipital electrode site) so as to collect SSVEP from the primary visual cortex \cite{norciaSteadystateVisualEvoked2015}. The EEG also has one reference electrode above the left ear, M1 position, and one ground electrode above the right ear, M2 position. \\
An audio cable leads the 3 pole signal into a amplifier before being imported into the microphone port of a computer which works as EEG recorder. \\
%
{\bf{Stimulus setup.}}
Modulated light intensities are provided either by a digital light projector or by a standard LCD computer screen. When using the screen e.g. to calibrate the SSVEP read-out, the screen displays uniform grayscale intensities with state of `on' and `off' as a flickering stimulus and with a square waveform,  Fig.~\ref{Screen_EEG_sys}. The subject/viewer wearing the EEG headset observes the screen flashing at a fixed frequency ($f=1/T$) and duty cycle (50\%).
There are two main features that need to be taken into account: one is the frame rate ($F_{rate}$) of the screen expressed in frames per second or FPS, the other is the bit depth of the image. Normally, computer screens operate at 60 Hz frame rate with an 8-bit intensity range. So a flashing or flickering image frequency can be achieved, ranging from 0 to 30Hz and the relative intensity can range from 0 to 255 bits. Mathematically, the various parameters and flicker rate can be expressed as:
\begin{equation}\label{Screen_flashing_expression}
\begin{split}
& I_{screen}[j]=A\times Square(2 \pi f t[j],Duty);\\
& A\in \{0,1,2,3,...,255\};\\
& f=[3,F_{rate}/2);\\
& Duty=(0\%, 100\%);\\
& t[j]=j/F_{rate};j \in \mathbb{W};
\end{split}
\end{equation}
where $A$ is the illumination intensity, $f$ is the flicker frequency, and $Duty$ is the duty cycle. $Square()$ is the square wave function and $t[j]$ is the time stamp of each frame.\\
%
%
%
{\bf{The SSVEP  energy heatmap and calibration with three  different people.}}
\begin{figure}[t]%
\centering
\includegraphics[width=\columnwidth]{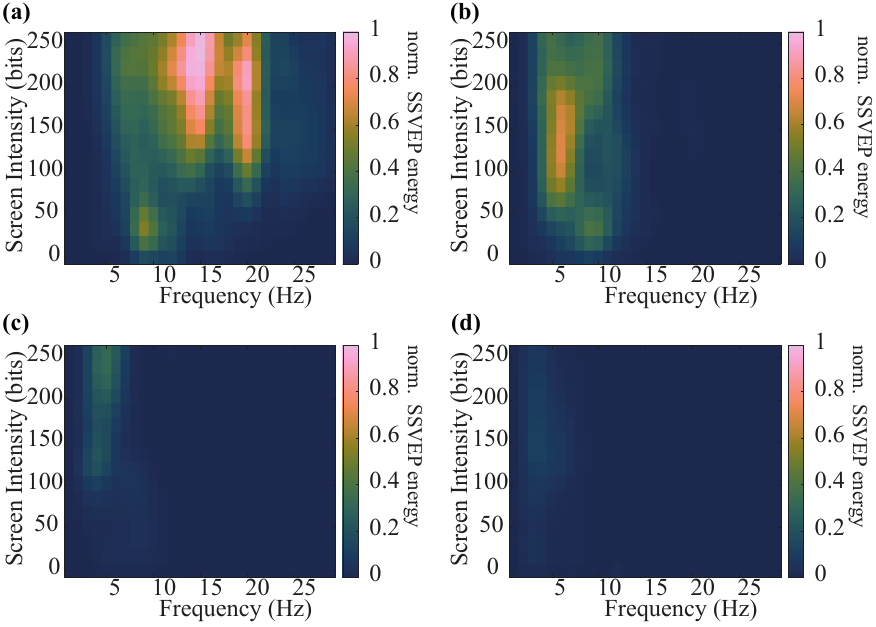}
\caption{Measured SSVEP Harmonic heat maps for varying light modulation frequency and illumination for the, {(a)} fundamental, {(b)} second  harmonic, {(c)} third harmonic  and, {(d)} fourth harmonic. The SSVEP signals in all four figures are normalised to the maximum recorded value for the fundamental signal in {\bf{a}}. 
}\label{F_I_harmonic_map}
\end{figure}
{ The calibration curves from different people are shown in Fig.~\ref{I_vs_H_curve_other}.  All three graphs [(a), (b) and (c) refer to subjects 1, 2 and 3] have in common a series of features that despite the variation in terms of quantitative values, still allow to define some general operating rules for computational ghost imaging with the brain. Specifically, all three show that at 15 Hz, the SSVEP increases monotonically for the intensity range used here. For 6 Hz modulation, all three SSVEP show a general trend to increase up to about one third of the total illumination intensity and then show a decrease or non-monotonic dependence. For example, subject 2, (b), shows a monotonic increase up to higher light intensities. Our conclusion is that a conservative approach of limiting light intensity to values corresponding to $\sim50$ Lumens 
should guarantee a monotonic SSVEP dependence in most individuals. A systematic study of this across a more statistically relevant pool of individuals was outside the scope of the current work but will be investigated in future studies. \\}
\begin{figure}[t]
	\centering
	\includegraphics[width=\columnwidth]{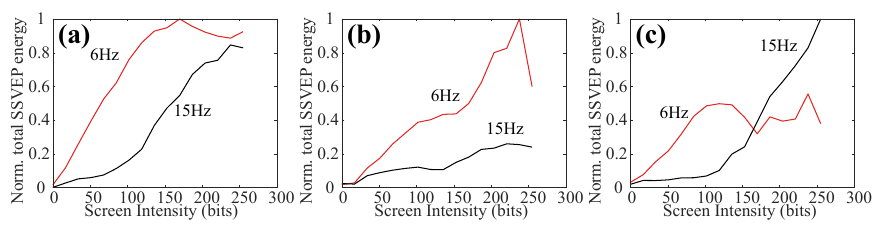}
	\caption{Measured total SSVEP energy at 6 Hz and 15 Hz from three subjects. }\label{I_vs_H_curve_other}
\end{figure}
{\bf{Tile-based intensity modulation.}}
\begin{figure}[t]
	\centering
	\includegraphics[width=\columnwidth]{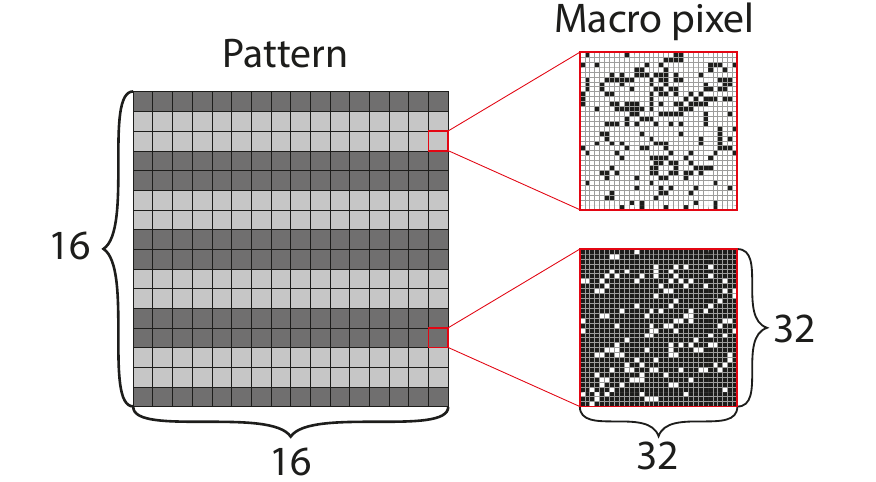}
	\caption{Tile macro-pixel modulation for the digital light projector (DLP). Each macro pixel in the $16 \times 16$ image array consists of $32 \times 32$ physical pixels in the DLP. The total number of physical pixels in each macro-pixel that are switched on, will determine the total luminosity of the macro-pixel with a range from 0 (no p[ixels on) to 1024 (all pixels on). Once the intensity and thus the total number of on-pixels is decided, these pixels are then chosen with a random pattern within the 32x32 macro-pixel array so as to avoid any possible aliasing or unwanted light patterning due to structure within the macro-pixel. }\label{Tile_modulation}
\end{figure}
For a DLP based projector, 
the Pulse-width modulation (PWM) is typically used to adjust the intensity in  8-bit images, which will affect the time-domain waveform of the stimulus. 
So, a different method has been proposed in this work, i.e.  `Tile-based intensity modulation', which achieves intensity modulation at the cost of resolution. As shown in Fig.~\ref{Tile_modulation}, each macro-pixel (or `tile') in a $16 \times 16$ pattern (i.e. the main image) is composed of $32 \times 32$ sub-pixels.
%
The intensity value of each tile macro-pixel can be achieved  by changing the percentage ($p$) of pixels actually switched on in each tile macro-pixel.  For example, when $p=0.5$, half of total tile pixels is selected randomly and lighted.\\
%
%
Thus for example, as shown in the Fig.~\ref{Tile_modulation}, the number of intensity values that a single tile macro pixel can achieve is 1025 (from 0 to 1024).\\
%
%
%
{\bf{Calibration of GI with the brain.}}
When using a digital light projector for actual ghost imaging, the same setup is used for both calibration and illumination of a target object. As described above, we vary the projected light intensity 
and monitor the output (SSVEP intensity) in order to verify a monotonic or close-to-linear response for the various objects we wish to illuminate. 
%
%
During the calibration experiment, the projector is controlled to flash at a fixed flashing frequency with varying intensity, with illumination times of 4 seconds with a short 0.5 second pause in between each illumination period.  The SSVEP is recorded for each illumination time.
The resulting calibration curve (In-Out intensity curve) 
is shown in Fig. \ref{GI_sys_calibration_curve}.\\
%
A linear range is found for $0<P\lesssim0.3$. 
This linear range  then guides the design of the patterns, which is used to re-scale and bias the binary patterns in order to ensure that the SSVEP is acquired in the linear range.\\
%
%
We also define a threshold that is used to distinguish whether the pattern has any overlap with object. The threshold value in Fig.~\ref{GI_sys_calibration_curve}  is set to $4\times10^{-4}$ and is determined by an empirical relationship that is based on our data and is described in more detail below.\\
\begin{figure}[t]%
	\centering
	\includegraphics[width=8cm]{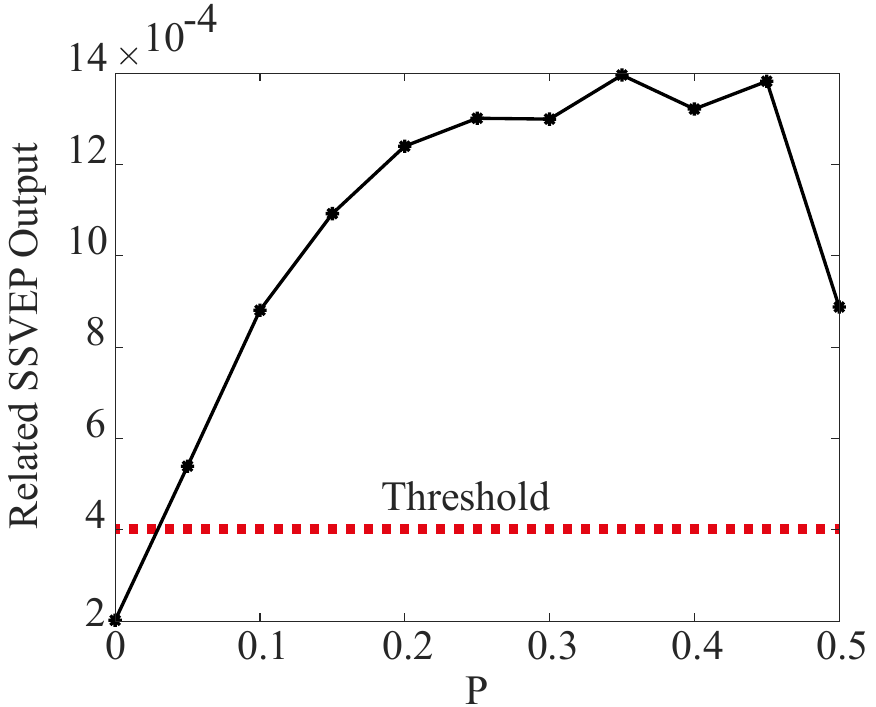}
	\caption{Total SSVEP energy (measured in arb. units) that is used as a calibration curve for neurofeedback ghost imaging. The red dashed line indicates the threshold at which we attributed signals to be consistently above the noise level. This threshold is used in the Hadamard carving step to distinguish `zero' SSVEP output (bucket) from non-zero values. 
	} \label{GI_sys_calibration_curve}
\end{figure}
{\bf{Image segmentation.}}
In standard GI, image resolution is determined by the number of patterns that are used to illuminate the object. If $t$ is the acquisition time for each pattern, then the total acquisition time is $T\propto N\times t$. The acquisition time $t$ will also determine the overall signal to noise ratio (SNR) in the final image. Under the assumption of a fixed light source intensity and illumination area,  increasing the number of pixels $N$ will also lead to a corresponding decrease in signal, i.e. the number of photons collected per pixel and hence SNR will scale inversely as N. This can be compensated for by increasing $t$ proportionally to $N$. Therefore, in order to maintain a fixed SNR as the number of pixels is varied, the total acquisition time is $T\propto N^2$. A detailed description of this is also give by Johnson et al. \cite{padgett2020}.

This therefore implies that there is an advantage to segmenting a higher resolution image into $q$ smaller sub-images of size $N/q$. Indeed, a fixed SNR can now be obtained with a total acquisition time $T\propto q\times(N/q)^2=N^2/q$.\\
{Therefore, in all our measurements, the object is illuminated only one column or stripe at a time, i.e. we project the Hadamard matrix column vectors as single-line elements that illuminate the object at a horizontal pixel location, starting e.g. from the left of the object. Once all of the patterns from (the carved) $H_c$ have been projected at this fixed horizontal location, we then shift one pixel to the right and repeat the projection and carving sequence. This procedure is iteratively repeated until the object has been fully scanned. }\\
{\bf{Hadamard matrices and image reconstruction.}}
The Hadamard matrices, $H_m$ are $2^m \times 2^m$ matrices that can be defined recursively:
\begin{equation}\label{hadamard_expression}
	\begin{split}
		{\displaystyle H_{m}={\frac {1}{\sqrt {2}}}{\begin{pmatrix}H_{m-1}&H_{m-1}\\H_{m-1}&-H_{m-1}\end{pmatrix}}};H_{0} = 1;
	\end{split}
\end{equation}
Hadamard patterns used in the GI typically consist of 0 and 1 values (as light intensities cannot take on negative values) so one typically uses $\hat{H}_{m}$ matrices that are defined as ${H}_{m}$ zeros replacing any negative values. 
The bucket value $B$ measured by a standard bucket detector (or equivalently, in our work, by the human visual system) can be expressed as: 
\begin{equation}\label{hadamard_GI_expression}
	\begin{split}
		B_{N \times 1}=\hat{H}^T_{N \times N} \cdot O_{N \times 1};
	\end{split}
\end{equation}
where 
$\hat{H}^T_{N \times N}$ is the transpose of the Hadamard pattern matrix and $\hat{H}_{N \times N}$ consist of N column vectors (patterns), namely, $\hat{H}_{N \times N}=[\hat{H}_{1},\hat{H}_{2}, \dots, \hat{H}_{N}]$.
Equation~\ref{hadamard_GI_expression} can be inverted to reconstruct the object $O$ from the measured bucket values by multiplying to the left and right by $\hat{H}_{N\times N}$ and then inverting to obtain
\begin{equation}\label{hadamard_GI_expression2}
	\begin{split}
		O_{N \times 1}=(\hat{H}_{N\times N} \cdot \hat{H}^T_{N \times N})^{-1}  \hat{H}_{N\times N} \cdot B_{N \times 1}.
	\end{split}
\end{equation}
If one uses the correctly defined Hadamard patterns with values -1 and 1, then $(\hat{H}_{N\times N} \cdot \hat{H}^T_{N \times N})^{-1}$ is just the identity matrix and one recovers the standard formula, i.e. $O=\sum a_nH_n$, where $a_n$ are the individual values of the bucket vector, $B$. However, as mentioned above, we do not measure negative intensity values and the Hadamard matrix is redefined by putting negative values to zero. We therefore use the more general formula Eq.~\ref{hadamard_GI_expression2} for reconstruction of images. This is also the formula provided in the main text.\\
{\bf{Hadamard matrix carving.}}
In standard GI, $N$ measurements (patterns) are required. However, this does not scale favourably with increasing pixel count, in particular for the total acquisition time that scales with $N^2$, as discussed above. We therefore introduce a "Hadamard matrix carving" method to reduce the imaging time by carving (i.e. reducing) the pattern matrix dynamically during the measurement phase.\\
Importantly, no prior knowledge of the object itself is required in order to establish how to select the required illumination patterns. In general, the more sparse the object is, the larger $p$ will be and the shorter the total acquisition time, i.e. the technique self-adapts to the object itself, always optimising the number of required patterns according the scene.\\
The starting point is the standard Hadamard matrix, $H$, with $N$ patterns. This is transformed to $H_c=RHC$, where the `row carving matrix', $R$,  and the `column carving matrix', $C$, are diagonal matrices:
\begin{equation}\label{Carving matrix}
	\begin{split}
		& R_{i,i}=U(-H_{i,j});\\
		& C_{i,i}=
		\begin{cases}
			0&r(i) \leq r(i-1)\\
			1&otherwise\\
		\end{cases}\\
		& r(i)=Rank((RH)_{:,1:i});
	\end{split}
\end{equation}
where $r(i)$ stand for the rank of the first $i$ columns of the matrix and $U$ is the unit step function: ${\displaystyle U(x)=\mathbf {1} _{x>0}}$. 

\begin{figure}[t]%
\centering
\includegraphics[width=\columnwidth]{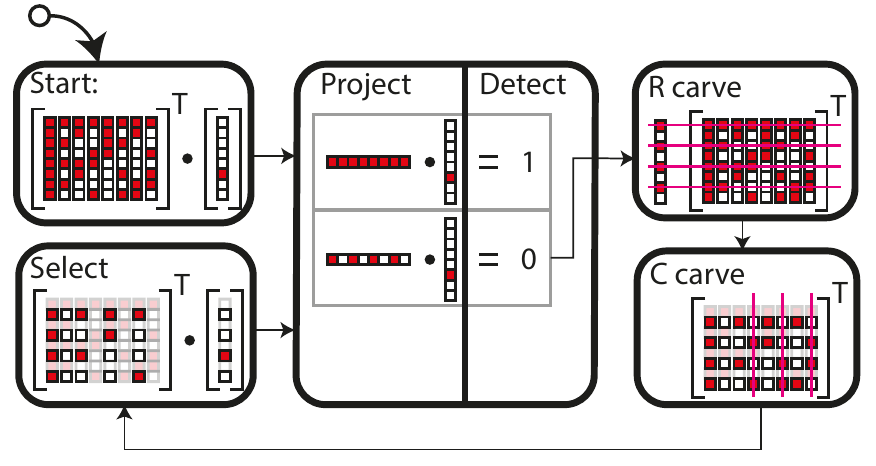}
\caption{Flow chart with a worked out example of Hadamard carving applied to the case $N=8$. 
}\label{Hadamard_carving_flow_}
\end{figure}

Beyond the formal definitions, the operations can be understood by following the work flow of the Hadamard matrix carving, shown in Fig.~\ref{Hadamard_carving_flow_}. This example explicitly works out, step by step, an example for the case $N=8$.\\
\begin{figure}[t]%
	\centering
	\includegraphics[width=\columnwidth]{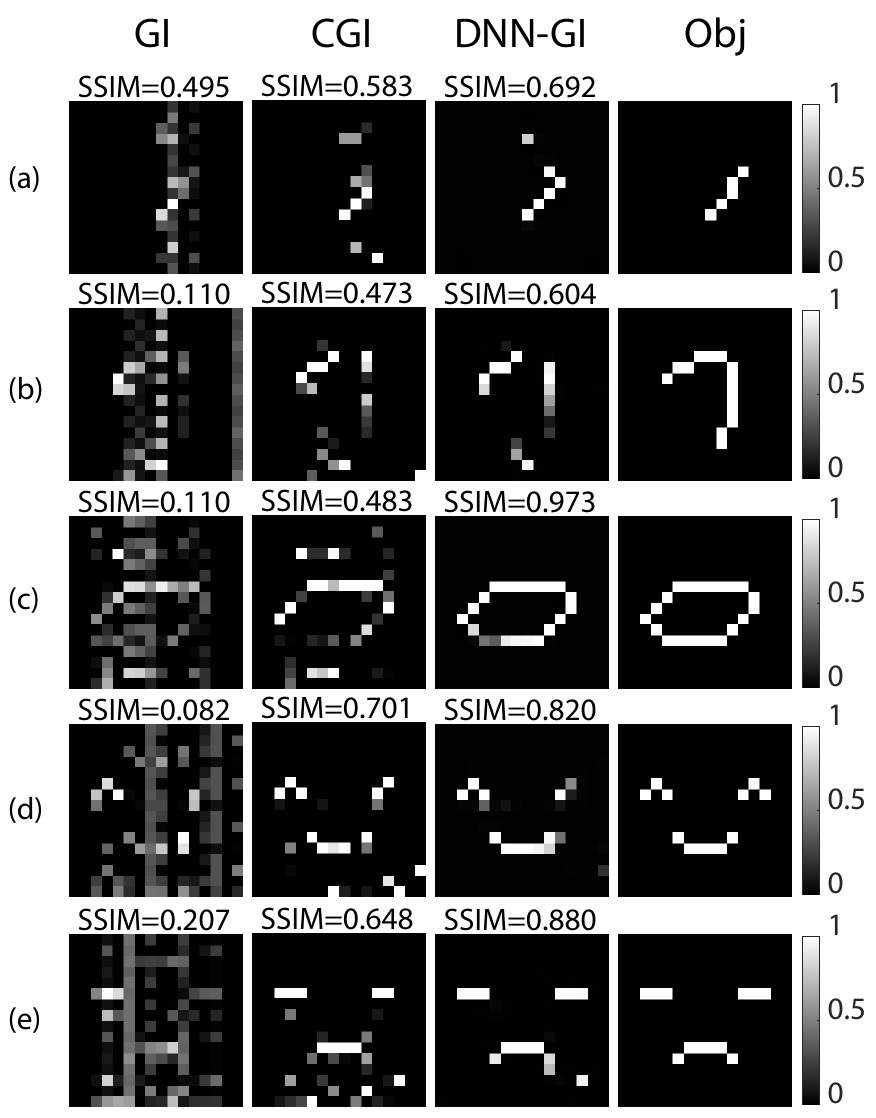}
	\caption{Neurofeedback ghost imaging results.  The four columns indicate results (from left to right) using the standard GI approach, carved GI (CGI), DNN GI reconstruction and the ground truth object.  Row (a) is for the number ``1'' with 42 projected patterns; (b) for the number ``7'' with 60 patterns; Row (c) for the number ``0'' with 87 patterns; Row (d) for a smiley face with 74 patterns; and Row (e) for a sad face with 76 patterns.
	}\label{ALl_imaging_results_SM}
\end{figure}
The standard $H$ matrix, each column consisting of an unwrapped Hadamard pattern (adjusted as explained above such that negative values are replaced with 0)  is applied to the object (unwrapped single column vector). We proceed to `project' the first column of $H$ onto the object. The row-column multiplication gives a number whose value indicates the overlap between the given Hadamard pattern and the object. In an experiment, this will correspond to the measured bucket value either by a photodiode or, in our case, by the eye and then read out as an SSVEP. If this detection value is finite (i.e. there is overlap between the pattern and the object), we keep the associated pattern. In the shown example, the second column (transposed) and multiplied by the object gives back a zero, i.e. there is no overlap between the pattern and the object. We therefore proceed to `carve' this pattern out of $H$. This is achieved by first performing a `row carving', i.e. all row elements corresponding to row numbers in which the pattern has value 1, are removed. The result matrix $RH$ is no longer square and has redundant column vectors, i.e. $RH$ can be reduced to a square matrix of rank $N/2$ by applying a `column carving', $C$. The $C$ operation can be understood as taking the columns, from left to right and keeping only those columns that increase the rank of the matrix. When adding an additional column to the right no longer increases the rank, this column can be removed. For the example shown in Fig.~\ref{Hadamard_carving_flow_}, the final rank of $H_c=RHC$ is $N/2=4$ after removing columns 4, 6 and 8.\\
At this point, the carved Hadamard matrix $H_c$ is applied to again to the object and the process is repeated. In the specific example show here, this will lead to a second carving process and the final carved $H_c$ will have rank $N/4=2$.\\
As can be seen from this example, the total reduction factor scales as $2^p$, as each carving step results in rank reduction of a factor 2 and $p$ will depend specifically on the details of the object. For example, a uniform object that extends over the full filed of view will have an overlap with all Hadamard patterns, i.e. $p=0$ and there will be no advantage. At the opposite extreme, very sparse objects, e.g. a single pixel in the field of view, will have no overlap with most patterns, thus maximising the value of $p$. As can be seen in the results shown in the main manuscript, this carving approach typically leads to reductions in pattern numbers and therefore also in total acquisition time in the order of $\sim73\%$.\\
In practice, all SSVEP measurements will be affected by noise so we need to apply a threshold in order to separate out bucket values that are zero (within the noise limit) and values that are greater than zero. This threshold is the same as shown above that is determined during the `calibration' process.\\
\textbf{Noise distribution of the SSVEP energy. }
The Hadamard carving procedure requires determining if a pattern has a zero overlap with the object, i.e. if the measured intensity is also zero. However, in the presence of noise, we must instead determine a threshold value, below which we consider the measured signal to be zero. We therefore first characterised the actual noise present in the SSVEP measurements.\\
We repeated the calibration with the LCD screen 30 times. The average of the SSVEP read-out, $\mu$, increases with light intensity up to around 150 bit (half maximum intensity) illumination, in agreement with Fig.~2(b) in the main text.
We also find that thee noise standard deviation scales as $\sigma\sim0.4\mu$. We therefore use this scaling to model noise for the DNN synthetic data (see below). \\
{\textbf{Threshold relation. }
The threshold value for the experiments that is used to distinguish between "zero" and "non-zero" SSVEP output is given by $\mu_0 + 2sigma$, where $\mu_0$ is the SSVEP with zero illumination intensity and we use $\sigma=0.5 \mu_0$.\\
\begin{figure}[t]%
\centering
\includegraphics[width=\columnwidth]{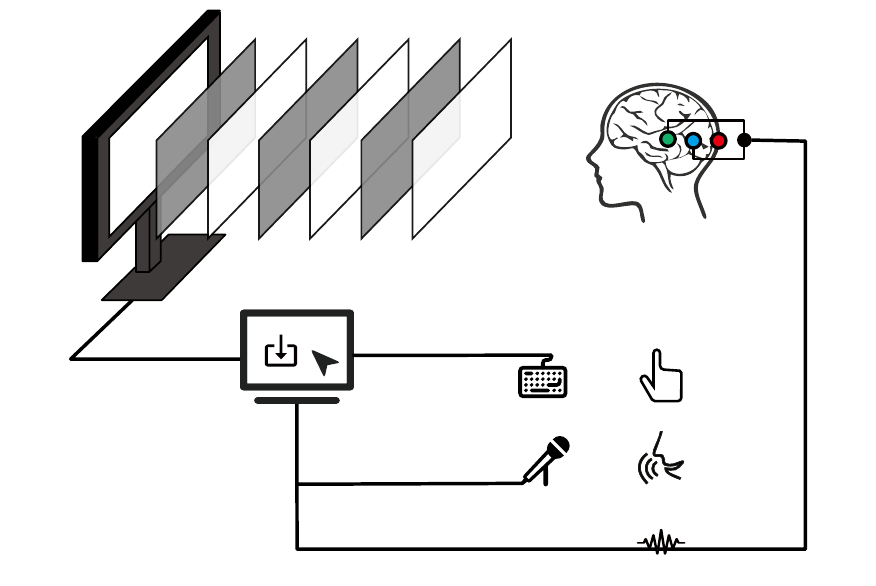}
\caption{conscious/nonconscious (explicit/non explicit) setup with LCD screen stimulus. The human actuator and the human EEG will be recorded as the response.} \label{Screen_EEG_sys_subconscious_setup}
\end{figure}
{\bf{Deep Neural Network ghost image (DNN-GI) reconstruction.}}
We implement an end-to-end deep neural network (DNN) that is the denoising DNN (``denoisingnetwork'' DnCCN) implemented in Matlab and based on Ref.~\cite{zhang2017beyond}. The DNN is a CNN that consists  of 59 layers with full details that can be found in Ref.~\cite{zhang2017beyond}. We note that the same network, although originally demonstrated only for denoising purposes, i.e. with a noisy image at the input and a denoised output image, we successfully used the same architecture to take at the input a 16x16 matrix of SSVEP total energy values and output a denoised image.\\
Training is performed using synthetic data based on  MNIST digit images are resized from  $28 \times 28$ into $16 \times 16 $ images (to have the same size as the images in our experiments), and are then used to generate the bucket values via CGI simulated on a computer, followed by adding noise using the noise model described above.} \\
Examples of reconstructed images are shown in Fig.~\ref{ALl_imaging_results_SM}. Rows (a) and (b) show examples of hand-written digits (a 1 and a 7, respectively). Rows (c), (d) and (e) are the same results as shown in the main text and demonstrate how the imaging is capable of generalising beyond the MNIST digits used for the training of the DNN.\\
{
{\bf{Conscious/nonconscious measurement protocol.}}
The conscious explicit response is the human actuator via keyboard or microphone while nonconscious non-explicit response is the EEG read-out via EEG headset.

The setup is shown in Fig.~\ref{Screen_EEG_sys_subconscious_setup}. Instead of using a sweep intensity of the 6Hz flickering stimulus, the intensities is equal to the bucket values from the ghost imaging simulation, Equation~\ref{hadamard_GI_expression}. In the simulation, the 8-by-8 object (digit “7”) is used and there are 64 intensities. The subject/viewer wearing the EEG headset observes the screen and was asked to do the tasks (EEG-only, EEG-Keyboard, and EEG-Mic).
In the EEG-only experiment, each flickering stimulus will last for 2s with a 0.5s interval time.
However, the EEG-Keyboard experiment will ask the subject to type the perceived intensity in numbers ranging from 0 to 15, while speaking (in the extra 2s) is implemented in the EEG-Mic experiment. The speaking voice was recorded and transcribed into the number (from 0 to 15) after the experiment.
\begin{figure}[t]%
	\centering
	\includegraphics[width=\columnwidth]{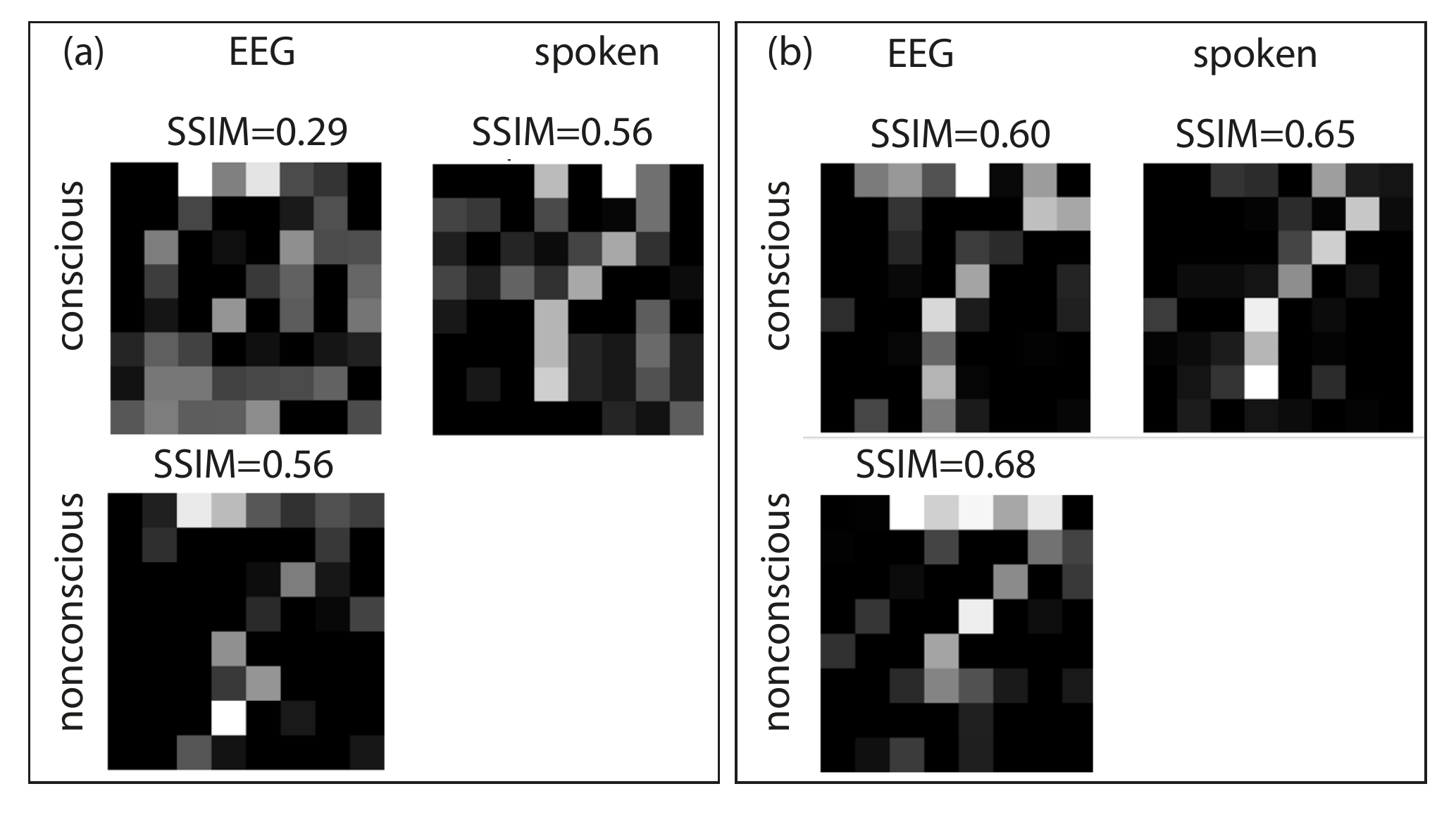}
	\caption{\label{sub2}  {{Nonconscious versus conscious (explicit) ghost imaging: (a) and (b) show the same experiment, repeated by two different subjects. Grayscale values corresponding to simulated ghost imaging bucket intensity values for the digit ``7'' are projected on a screen that is observed by the subject. Each experiment is then composed of two sessions. In one session (labelled as ``nonconscious''), only the EEG readout is used to reconstruct the final ghost image. In the second session (labelled as ``conscious''), the subject is asked to verbally communicate a numerical value between 0 and 15 that represents their estimate of the projected intensity. This is performed in parallel with an EEG recording. We then compare the ghost image retrieval using the verbally communicated values and the EEG readout. The ``nonconscious''  EEG ghost image and the ``conscious'' verbally communicated-values ghost image correspond well and have very similar SSIM. However, for both subjects, the EEG read-out ghost image deteriorates considerably in the ``coscious'' case, in agreement with other measurements shown in the main paper.}} } 
\end{figure}
The number from the human actuator and the harmonic sum of EEG-only read out will be regarded as the bucket values reconstructing the image of the object and estimating the conscious/nonconscious response's SSIM. Results of these experiments are shown in the main paper for the case in which the bucket values are typed on a keyboard. Figure~\ref{sub2} show additional results where we perform the same test with two different subjects, now verbally communicating the intensity values and compare these to EEG read-out ghost imaging. Similarly to the results shown in the main paper, the explicit readout appears to interfere with the nonconscious EEG readout.

}

\end{document}